\documentclass[10pt,journal,compsoc]{IEEEtran}
\pdfoutput=1

\ifCLASSOPTIONcompsoc
  \usepackage[nocompress]{cite}
\else
  \usepackage{cite}
\fi
\ifCLASSINFOpdf
   \usepackage[pdftex]{graphicx}
   \graphicspath{{./figures/}}
\else
\fi
\usepackage{amsmath}
\usepackage{array}

\usepackage{tikz}
\usepackage{amsmath}
\usepackage{amssymb}
\usepackage{multicol,multirow}
\usepackage{color}
\usepackage{pgfplots}

\begin{document}
%
% paper title
% Titles are generally capitalized except for words such as a, an, and, as,
% at, but, by, for, in, nor, of, on, or, the, to and up, which are usually
% not capitalized unless they are the first or last word of the title.
% Linebreaks \\ can be used within to get better formatting as desired.
% Do not put math or special symbols in the title.
\title{Learning Occlusion-Aware View Synthesis \\for Light Fields}
%
%
% author names and IEEE memberships
% note positions of commas and nonbreaking spaces ( ~ ) LaTeX will not break
% a structure at a ~ so this keeps an author's name from being broken across
% two lines.
% use \thanks{} to gain access to the first footnote area
% a separate \thanks must be used for each paragraph as LaTeX2e's \thanks
% was not built to handle multiple paragraphs
%
%
%\IEEEcompsocitemizethanks is a special \thanks that produces the bulleted
% lists the Computer Society journals use for "first footnote" author
% affiliations. Use \IEEEcompsocthanksitem which works much like \item
% for each affiliation group. When not in compsoc mode,
% \IEEEcompsocitemizethanks becomes like \thanks and
% \IEEEcompsocthanksitem becomes a line break with idention. This
% facilitates dual compilation, although admittedly the differences in the
% desired content of \author between the different types of papers makes a
% one-size-fits-all approach a daunting prospect. For instance, compsoc 
% journal papers have the author affiliations above the "Manuscript
% received ..."  text while in non-compsoc journals this is reversed. Sigh.

\author{Julia~Navarro~and~Neus~Sabater% <-this % stops a space
\IEEEcompsocitemizethanks{\IEEEcompsocthanksitem J. Navarro is with Universitat de les Illes Balears, Palma, 07122 Spain. \protect\\ E-mail: julia.navarro@uib.es}% <-this % stops a space
\IEEEcompsocitemizethanks{\IEEEcompsocthanksitem N. Sabater is with Technicolor R\&I, Cesson-S{\'e}vign{\'e}, 35576 France. \protect\\ E-mail: neus.sabater@technicolor.com}% <-this % stops a space
%\thanks{Manuscript received April 19, 2005; revised August 26, 2015.}
}

\IEEEtitleabstractindextext{%
\begin{abstract}
In this work, we present a novel learning-based approach to synthesize new views of a light field image. 
In particular, given the four corner views of a light field, the presented method estimates any in-between view. 
We use three sequential convolutional neural networks for feature extraction, scene geometry estimation and view selection. Compared to state-of-the-art approaches, in order to handle occlusions we propose to estimate a different disparity map per view. Jointly with the view selection network, this strategy shows to be the most important to have proper reconstructions near object boundaries. Ablation studies and comparison against the state of the art on Lytro light fields show the superior performance of the proposed method. Furthermore, the method is adapted and tested on light fields with wide baselines acquired with a camera array and, in spite of having to deal with large occluded areas, the proposed approach yields very promising results. 
\end{abstract}

% Note that keywords are not normally used for peerreview papers.
}

% make the title area
\maketitle

% To allow for easy dual compilation without having to reenter the
% abstract/keywords data, the \IEEEtitleabstractindextext text will
% not be used in maketitle, but will appear (i.e., to be "transported")
% here as \IEEEdisplaynontitleabstractindextext when the compsoc 
% or transmag modes are not selected <OR> if conference mode is selected 
% - because all conference papers position the abstract like regular
% papers do.
\IEEEdisplaynontitleabstractindextext
% \IEEEdisplaynontitleabstractindextext has no effect when using
% compsoc or transmag under a non-conference mode.

% For peer review papers, you can put extra information on the cover
% page as needed:
% \ifCLASSOPTIONpeerreview
% \begin{center} \bfseries EDICS Category: 3-BBND \end{center}
% \fi
%
% For peerreview papers, this IEEEtran command inserts a page break and
% creates the second title. It will be ignored for other modes.
\IEEEpeerreviewmaketitle

%%%%%%%%%%%%%%%%%%%%%%%%%%%%%%%%%%%%%%%%%%
%%															  %%
%%						INTRODUCTION				        	           %%
%%														 	  %%
%%%%%%%%%%%%%%%%%%%%%%%%%%%%%%%%%%%%%%%%%%
\IEEEraisesectionheading{\section{Introduction}\label{sec:introduction}}
% Computer Society journal (but not conference!) papers do something unusual
% with the very first section heading (almost always called "Introduction").
% They place it ABOVE the main text! IEEEtran.cls does not automatically do
% this for you, but you can achieve this effect with the provided
% \IEEEraisesectionheading{} command. Note the need to keep any \label that
% is to refer to the section immediately after \section in the above as
% \IEEEraisesectionheading puts \section within a raised box.

% The very first letter is a 2 line initial drop letter followed
% by the rest of the first word in caps (small caps for compsoc).
% 
% form to use if the first word consists of a single letter:
% \IEEEPARstart{A}{demo} file is ....
% 
% form to use if you need the single drop letter followed by
% normal text (unknown if ever used by the IEEE):
% \IEEEPARstart{A}{}demo file is ....
% 
% Some journals put the first two words in caps:
% \IEEEPARstart{T}{his demo} file is ....
% 
% Here we have the typical use of a "T" for an initial drop letter
% and "HIS" in caps to complete the first word.
\IEEEPARstart{L}{ight} field imaging has recently gained importance due to the additional information that provides of the scene. 
Contrary to conventional 2D images that at each point capture the sum of all light rays coming from different angles, the 4D light field image captures the whole light information. A light field image can be considered as a collection of 2D images taken from different viewpoints that are arranged on a regular grid.

Plenoptic cameras such as Lytro~\cite{lytro} or camera arrays~\cite{sabater2017dataset,dkabala2016efficient} are among the different devices that can be used for the acquisition of these images. In the first case, given that the sensor resolution is limited, the additional information given from the different viewpoints comes at the cost of an important decrease in spatial resolution, compared to traditional cameras. Plenoptic cameras usually offer high angular resolutions ($14\times14$ views for Lytro Illum) with small baselines. On the other hand, camera arrays do not suffer from low spatial resolution but capturing a large number of views would be costly, and generally they capture sparse light fields with wide baselines. In addition, current smartphones also capture light fields using several cameras. However, they cannot provide high angular resolutions since it is not possible to have a large number of cameras in a cellphone.

Then, it is interesting to study the problem of new view synthesis for light fields. That is, the generation of images from novel viewpoints. With new view synthesis methods plenoptic cameras could be built to capture light fields with smaller angular resolution and thus provide higher spatial resolutions. Also, camera arrays and cellphones could increase the number of views using  view synthesis techniques. Besides, the generation of novel views would permit to navigate smoothly between the different images and be used for applications such as virtual reality~\cite{overbeck2018system}.

Over the last years, deep learning has had a great success in computer vision and image processing tasks.  Indeed, deep learning has proved to be competitive with respect to traditional approaches for different problems such as stereo~\cite{kendall2017end,chang2018pyramid}, optical flow~\cite{dosovitskiy2015flownet}, denoising~\cite{burger2012image} and super-resolution~\cite{dong2014learning}.  Furthermore, it has recently been applied to light field images for super-resolution~\cite{yoon2015learning}, depth estimation~\cite{shin2018epinet} or separation into diffuse and specular intrinsic components~\cite{alperovich2018light}.

Inspired by recent work on new view synthesis using deep learning~\cite{flynn2016deepstereo,kalantari2016learning,srinivasan2017learning}, we propose a novel learning-based solution to synthesize views of a light field image. Particularly, given de four corner views, we reconstruct any view in between. The approach is designed for plenoptic light fields captured with the Lytro Illum camera. Moreover, we adapt the model for wide baselines and very promising results are obtained in the case of light fields captured with a camera rig, which have larger occluded regions. 

We divide the problem of view synthesis into feature extraction, disparity estimation and view selection and use three sequential convolutional neural networks. Disparity is estimated between the virtual novel view and each corner view. In contrast to the recent approach from Kalantari et al.~\cite{kalantari2016learning}, in order to handle occlusions we propose to estimate a different disparity map for each corner image. The selection network detects occluded parts and discards them to reconstruct the novel view. This results in accurate reconstructions near object boundaries and occlusions, while the method in~\cite{kalantari2016learning} produces blurred results at these regions. Srinivasan et al.~\cite{srinivasan2017learning} reconstruct the 4D light field from the center view. While their problem is more challenging than ours, they work with images with simple and similar geometry, being unable to deal with more complex scenes. In spite of having been trained on the same dataset, our approach performs properly on different scenes. Flynn et al.~\cite{flynn2016deepstereo} cope with wide baseline images by providing to the networks a plane sweep volume built from the input views. While this is memory and time consuming (it takes minutes to synthesize a novel view), our method takes few seconds to predict the novel image directly from the input views.

Although the purpose of the work is view synthesis, the presented method is able to estimate disparity. Since we only need a large collection of light fields for training and any ground truth depth is needed, it learns disparity in an unsupervised manner. Furthermore, the disparity estimation carried out by our method is competitive with respect to the state of the art.

The work is organized as follows. Section \ref{sec:previouswork} reviews the state of the art on view synthesis. In Section \ref{sec:proposed}, we present the proposed new view synthesis model. In Section \ref{sec:experiments} we evaluate the presented method with extensive experiments and comparisons. Finally, Section \ref{sec:conclusions} concludes the paper.

%%%%%%%%%%%%%%%%%%%%%%%%%%%%%%%%%%%%%%%%%%
%%															  %%
%%						PREVIOUS WORK				        	           %%
%%														 	  %%
%%%%%%%%%%%%%%%%%%%%%%%%%%%%%%%%%%%%%%%%%%
\section{Previous Work}\label{sec:previouswork}

In this section, we review the state of the art on new view synthesis. We divide these methods into non-learning and learning-based approaches. In addition, as the problem is closely related to video frame interpolation, we also introduce recent methods tackling this problem. 

\subsection{Non-Learning Approaches}
Traditional methods generate novel views of scenes and objects from an arbitrary collection of images from the scene, which is known as image-based rendering (IBR). Generally, these methods first predict the scene geometry and then generate the novel image from the warped views. Chaurasia et al.~\cite{chaurasia2013depth} proposed a depth-synthesis approach using graphs operating over an over-segmentation of the input views. Goesele et al.~\cite{goesele2010ambient} introduced ambient point clouds to represent areas with uncertain depth. Other methods estimate the novel image without explicitly estimating geometry. Fitzgibbon et al.~\cite{fitzgibbon2005image} avoided the explicit depth computation and used image-based priors. Shechtman et al.~\cite{shechtman2010regenerative} used a patch-based optimization framework.

Among the non-learning view synthesis methods for light field images we find the variational model from 
Wanner and Goldluecke~\cite{wanner2014variational}. Given the disparity maps at the input views, the energy functional penalizes deviations between each warped input onto the novel position and the unknown view. This term incorporates a mask to account for occlusions. Also, a smoothness term for the novel image using total variation is included. Shi et al.~\cite{shi2014light} work in the continuous Fourier domain to reconstruct dense light fields from a 1D set of viewpoints. Zhang et al.~\cite{zhang2015light} proposed a phase-based approach to reconstruct a whole light field from a stereo pair with disparities smaller than five pixels. Penner and Zhang~\cite{penner2017soft} generate novel views of plenoptic light fields and camera arrays by means of a soft 3D model of the scene geometry.

\subsection{Learning-based Methods}
More recent methods make use of convolutional neural networks (CNN) to model the problem. Yoon et al.~\cite{yoon2015learning} jointly model spatial and angular light field super resolution with a CNN. The spatially upsampled result is the input to the angular super resolution network, which consists of a single CNN. Wu et al.~\cite{wu2017light} reconstruct any view of the light field given a sparse set of views by using a CNN on epipolar plane images. Similarly, Wang et al.~\cite{wang2018end} combine 2D and 3D convolutions applied on epipolar plane images to reconstruct the entire light field. From the four corner views of a light field, Kalantari et al.~\cite{kalantari2016learning} propose two CNN to synthesize any view in between. They manually extract features by first warping the input images at different disparity levels and then computing mean and variance at each level. Given these features, the first network computes one disparity map for the unknown view, which is used to warp each corner image. The four warpings are combined through another CNN which outputs the predicted view. Srinivasan et al.~\cite{srinivasan2017learning} aim at reconstructing the whole light field given just the center view. A first network estimates a 4D depth map from the input image. These maps are used to warp the center view and obtain an initial estimate of the 4D light field, which is further refined through a residual network. The method is trained on images of flowers, all of them sharing similar geometry, and it fails when testing on more complex scenes.

Flynn et al.~\cite{flynn2016deepstereo} deal with wide-baseline images by building a plane sweep volume. This volume is the input to two different networks, one that outputs for each pixel and depth the probability of that pixel having that depth, and the other generates a color image at each depth plane. The point-wise product between probabilities and color images provide the novel view. A drawback of this approach is the need to build the plane sweep volume, which is memory and time consuming. Indeed, the authors have to generate images in small patches to save memory and it takes 12 minutes to synthesize a $512\times512$ image. Plane sweep volumes are also used by Zhou et al.~\cite{zhou2018stereo}, where the authors develop a method for view extrapolation given two images with small baseline, using an encoder-decoder architecture. Built upon this method, Srinivasan et al.~\cite{srinivasan2019pushing} further extend the possible lateral movement and improve reconstructions at disocclusions. The recent method from Mildenhall et al.~\cite{mildenhall2019local} also uses plane sweep volumes to synthesize novel views given an irregular grid of input images from the scene.

\subsection{Video Frame Interpolation}
Given two frames of a video, frame interpolation consists of predicting frames at novel time instants. Most approaches first estimate optical flow to warp the input frames to the target one and then proceed to combine these warpings. 

Liu et al.~\cite{liu2017video} proposed Deep Voxel Flow, a multiscale frame interpolation method. At three different scales, they compute the optical flow and a confidence map using encoder-decoder networks. The results from these scales are combined using two convolutional layers. The output vector field is used to warp the input images and both warpings are combined using the confidence map. Amersfoort et al.~\cite{van2017frame} estimate optical flow and confidence in a coarse-to-fine scheme to deal with large displacements. At each scale the method estimates a residual to refine both the flow estimation and confidence. The finest optical flow computation is at half resolution, and it is upsampled to warp the frames at full resolution and generate the novel one. This result is further refined through a CNN. Niklaus and Liu~\cite{niklaus2018context} use an existing deep learning method to compute forward and backward optical flows. These are used to warp the input frames as well as the features provided by the first layer of the ResNet18~\cite{he2016deep}. The four warpings are the input to a network with a GridNet architecture~\cite{fourure2017residual}. Jiang et al.~\cite{jiang2017super} use an encoder-decoder network to predict forward and backward flows. These are used to warp the frames to the desired time instant and input views, warpings and optical flows are introduced into another encoder-decoder network to refine the optical flow. This network outputs the refined flow, jointly with a confidence mask. Then, the input frames are warped with these refined flows and combined according to the confidences.

%%%%%%%%%%%%%%%%%%%%%%%%%%%% 
\begin{table}[t!]\centering
\caption{Networks architectures.}
\begin{tabular}{|c|cc@{\hskip 0.5em}cccc@{\hskip 0.5em}c@{\hskip 0.5em}|}\cline{2-8}
\multicolumn{1}{c|}{}& Name & k & r & In & Out & Act. f. & BN \\ \cline{2-8} \noalign{\smallskip}   \hline 

\multirow{11}{*}{\rotatebox{90}{Features CNN}}  &input && & & 5 & & \\ \cline{2-8}
&conv0 & $3\times3$ && 5 & 32 & ELU & $\checkmark$ \\
&conv1 & $3\times3$ && 32 & 32 & ELU & $\checkmark$ \\
&conv2 & $3\times3$ && 32 & 32 & ELU & $\checkmark$ \\
&conv3 & $3\times3$ && 32 & 32 & ELU & $\checkmark$ \\
&conv4 & $3\times3$ && 32 & 32 & ELU & $\checkmark$ \\ \cline{2-8}
&\multicolumn{7}{c|}{$\text{conv4}=\text{conv2}+\text{conv4}$} \\ \cline{2-8}
&pool0 & $16\times16$ && 32 & 32 & \multicolumn{2}{c|}{\multirow{2}{*}{avg. (conv4)}} \\
&pool1 & $8\times8$ && 32 & 32 & & \\ \cline{2-8}
&\multicolumn{7}{c|}{concatenate [conv2, conv4, pool0, pool1]}\\ \cline{2-8}
&conv5 & $3\times3$ && 128 & 32 & ELU & $\checkmark$ \\ \hline \noalign{\smallskip} \hline

\multirow{9}{*}{\rotatebox{90}{Disparity CNN}}  & input & & & & 130 & & \\  \cline{2-8}
&conv0 & $3\times3$ & 2 & 130 & 128 & ELU & $\checkmark$ \\
&conv1 & $3\times3$ & 4 & 128 & 128 & ELU & $\checkmark$ \\
&conv2 & $3\times3$ & 8 & 128 & 128 & ELU & $\checkmark$ \\
&conv3 & $3\times3$ & 16 & 128 & 128 & ELU & $\checkmark$ \\
&conv4 & $3\times3$ & & 128 & 64 & ELU & $\checkmark$ \\ 
&conv5 & $3\times3$ & & 64 & 64 & ELU & $\checkmark$ \\ 
&conv6 & $3\times3$ & & 64 & 4 & $\tanh$ &  \\  \cline{2-8}
&\multicolumn{7}{c|}{$d_{\text{max}}\cdot\text{conv6}$} \\ \hline \noalign{\smallskip}\hline

\multirow{9}{*}{\rotatebox{90}{Selection CNN}}& input & & &  & 18 & & \\ \cline{2-8}
&conv0 & $3\times3$  && 18 & 64 & ELU & $\checkmark$ \\
&conv1 & $3\times3$  && 64 & 128 & ELU & $\checkmark$ \\
&conv2 & $3\times3$  && 128 & 128 & ELU & $\checkmark$ \\
&conv3 & $3\times3$  && 128 & 128 & ELU & $\checkmark$ \\
&conv4 & $3\times3$   && 128 & 64 & ELU & $\checkmark$ \\ 
&conv5 & $3\times3$  && 64 & 32 & ELU & $\checkmark$ \\ 
&conv6 & $3\times3$  && 32 & 4 & $\tanh$ &  \\ \cline{2-8}
&\multicolumn{7}{c|}{Softmax with learned $\beta$} \\ \hline
\end{tabular}
\label{tab:networks}
\end{table}
%%%%%%%%%%%%%%%%%%%%%%%%%%%% 

%%%%%%%%%%%%%%%%%%%%%%%%%%%%%%%%%%%%%%%%%%
%%															  %%
%%						PROPOSED					        	           %%
%%														 	  %%
%%%%%%%%%%%%%%%%%%%%%%%%%%%%%%%%%%%%%%%%%%
\section{Proposed Method for View Synthesis}\label{sec:proposed}

Let $\Omega\subset\mathbb{R}^2$ be an open bounded domain, usually a rectangle in $\mathbb{R}^2$, and let us consider a light field image with $(N+1)\times(N+1)$ views, with $N\in\mathbb{N}, N\geq2$. Let us denote by $I_{p,q}:\Omega \rightarrow \mathbb{R}^3$ the view at the angular position $(p,q)$, with $p, q\in[0,N]$ and with the $(0,0)$ image being the one at the top-left corner. 

Given the four corner images $I_{0,0}, I_{0,N}, I_{N,0}$ and $I_{N,N}$ and the angular coordinates $(p,q)$ of any in-between view, the goal is to estimate the view $I_{p,q}$, That is, we aim at finding a function $f$ such that 
\begin{equation}
\hat{I}_{p,q} = f(p, q, I_{0,0}, I_{0,N}, I_{N,0}, I_{N,N}),
\end{equation}
with $\hat{I}_{p,q}$ being the estimated view at position $(p,q)$.

We model $f$ by using convolutional neural networks. One option would be to consider $f$ as a single network that from the four corner views and the coordinates of the novel position directly outputs the predicted view. However, as pointed out in~\cite{kalantari2016learning,srinivasan2017learning}, the relation between input and output is too complex to be modeled by just a single network. A proof of that is later shown in Section \ref{sec:experiments}.

\subsection{Proposed Model}
We split the problem into feature extraction, disparity estimation and view selection and use three different convolutional neural networks, one for each purpose. Features extracted from four input images are concatenated and used to estimate disparity. Then, input views are warped according to this disparity and four selection masks that will serve to perform a weighted average of the four warpings are estimated. 

The three networks additionally receive as input the coordinates $(p, q)$ of the novel view. In order to provide these coordinates to the convolutional networks, we consider images $P, Q:\Omega\rightarrow\mathbb{R}$ such that
\begin{equation}
\begin{split}
P(x,y) = p, \qquad \forall (x,y)\in\Omega, \\
Q(x,y) = q, \qquad \forall (x,y)\in\Omega. \\
\end{split}
\end{equation}
In the following we detail each stage of our algorithm.

\paragraph{Features CNN}
Compared to~\cite{kalantari2016learning} that extracts features manually, we use a convolutional neural network for this purpose. The features extraction network ($f_e$) is applied independently to each input image to compute a feature volume with 32 channels for each one of the four input views. These features should not depend on the image being processed and therefore weights are shared across all views. This network also receives as input the images $P$ and $Q$, which are concatenated to the considered image along the channel dimension, resulting in an input volume with 5 channels.

The network $f_e$ consists of a sequence of five convolutional layers with $3\times3$ kernels, including one residual block~\cite{he2016deep}. Average poolings with kernels $16\times16$ and $8\times8$ are then used to extract features at different scales, providing the network of more global information. Features from different layers are concatenated and finally fused with $3\times3$ convolutions. All convolutional layers are followed by an ELU activation and batch normalization~\cite{ioffe2015batch}. This architecture is a simplified version of the feature extraction stage proposed in~\cite{chang2018pyramid}. %See Table~\ref{tab:networks} for more details.

Let $F_{i,j}=f_e\left(P, Q, I_{i,j}\right)$ be the computed feature volume for image $I_{i,j}$, for $i,j\in\{0,N\}$. Then, the four volumes are concatenated,
\begin{equation}
\mathbf{F}=(F_{0,0}, F_{0,N}, F_{N,0}, F_{N,N}),
\end{equation}
and this 128-channel volume $\bf F$ is the input to the next stage.

\paragraph{Disparity CNN}
We assume that the views of the light field are arranged on a regular grid. Then, horizontal and vertical disparities are the same for consecutive views and thus the same estimated map is used in both components. For the same reason, disparities between each corner view and the virtual view are the same and one common map for the four images should be enough. In practice, however, the matching problem is not defined at occluded areas and, since occluded pixels are different depending on the view, it results in different disparity maps. Therefore, in contrast to~\cite{kalantari2016learning}, we let the network to estimate four different disparity maps $d_{i,j}$ depicting the displacement between $I_{i,j}$ and the virtual view $\hat{I}_{p,q}$, for $i,j\in\{0,N\}$. In Section \ref{sec:experiments} we show the advantages of using this strategy.

The disparity maps $\mathbf{d}=(d_{0,0}, d_{0,N}, d_{N,0}, d_{N,N})$ are computed from the angular position of the novel view and the four feature volumes, $\bf F$, through network $f_d$, 
\begin{equation}
\mathbf{d}=f_d(P, Q, \mathbf{F}).
\end{equation}
This network consists of seven convolutional layers, all of them with a filter size of $3\times3$. The first four ones use dilated convolutions at rates $2, 4, 8$ and $16$, respectively. The use of dilated convolutions permits to combine features at different resolutions and provide the network with more context. All layers but the last one use an ELU activation function and batch normalization. Last layer uses the hyperbolic tangent as activation function and no batch normalization is applied. The $\tanh$ rescales the output into the range $[-1,1]$. Then, the output disparity is multiplied by a constant $d_{\text{max}}$, which is the maximum allowed disparity magnitude. This way the output disparity will be in the range $[-d_{\text{max}}, d_{\text{max}}]$. For Lytro images, this value is set to $d_{max}=4$. %The detailed network is described in Table~\ref{tab:networks}.

\paragraph{Image Warping} 
The estimated disparity is used to warp each corner view in order to have them registered with the virtual one.
Let $I_{i,j}^w$ denote the warped image for view $I_{i,j}$. Then, for all $i,j\in\{0,N\}$,
\begin{equation}\label{eq:warpings}
I_{i,j}^w(x,y)= I_{i,j}(x+(i-p)d_{i,j}, y+(j-q)d_{i,j}),
\end{equation}
where $d_{i,j}$ is evaluated at pixel $(x,y)$. Warped images and disparity maps are concatenated to form the volume $\bf W$,
\begin{equation}
\mathbf{W} = (I_{0,0}^w, I_{0,N}^w, I_{N,0}^w, I_{N,N}^w).\end{equation}
This 12-channel volume $\bf W$, the depth maps $\mathbf{d}$ and images $P$ and $Q$ are the input to the selection network.

\paragraph{Selection CNN}
The task of the selection network ($f_s$) is to determine the contribution of each warped image $I_{i,j}^w$ to the final result. This will be achieved by computing four selection masks $(m_{0,0}, m_{0,N}, m_{N,0}, m_{N,N})=f_s(P, Q, \mathbf{W, d})$
such that $m_{i,j}(x,y)\in[0,1]$, for all $i,j\in\{0,N\}$, and
\begin{equation}
\sum_{i,j\in\{0,N\}} m_{i,j}(x,y) = 1, \qquad\forall(x,y)\in\Omega.
\end{equation}
Then, the predicted view is computed as a weighted average of the four warped images using as weights these selection masks,
\begin{equation}\label{eq:reconstruction}
\hat{I}_{p,q}(x,y)= \sum_{i,j\in\{0,N\}} m_{i,j}(x,y)\, I_{i,j}^w(x,y).
\end{equation}

The selection network $f_s$ consists of seven convolutional layers with $3\times3$ filters. All layers but the last one are followed by an ELU and batch normalization. At the last layer we use $\tanh$ and do not use batch normalization. Besides, at the last layer we also apply a softmax normalization along views,
\begin{equation}
\sigma_\beta(v_i(\mathbf{x})) = \frac{e^{\beta v_i(\mathbf{x})}}{\sum_{i=1}^4e^{\beta v_i(\mathbf{x})}}, \qquad \forall i\in\{1,2,3,4\},
\end{equation}
with $\mathbf{x}=(x,y)$ and $v_i$ being channel $i$ of the \texttt{conv6} layer. With the softmax we ensure that the sum of the selection weights over the four views equals one at each pixel. Moreover, we let the network to learn the parameter $\beta$. High values of this parameter encourage the network to select a single view, which is important at those areas that are only visible in one of the four images. The network has to be able to detect which regions of the novel view are also visible in the four corner ones. With these masks we discard inaccuracies in the warped images coming from occluded pixels. After training the network, the learned value is $\beta=8.01$.

%%%%%%%%%%%%%%%%%%%%%%%%%%%%% 
%\begin{figure*}[th!]\centering
%\includegraphics[width=\linewidth]{diagram/diagram.pdf}
%\caption{Diagram of the proposed new view synthesis method. We divide the problem into feature extraction, disparity estimation and view selection and use three different networks, one for each purpose. Features extracted from the four input images and the images containing the coordinates of the novel position $P$ and $Q$ are concatenated and used to estimate disparity. Then, input views are warped according to its corresponding disparity and we estimate four selection masks that are utilized to perform a weighted average of the four warpings.}
%\label{fig:diagram}
%\end{figure*}
%%%%%%%%%%%%%%%%%%%%%%%%%%%%% 
%
%Figure \ref{fig:diagram} illustrates the diagram of the proposed approach. 

Table~\ref{tab:networks} details the three presented networks. In the table, labels {\it In} and {\it Out} correspond to the number of channels of input and output volumes, respectively. BN denotes batch normalization~\cite{ioffe2015batch}, $k$ is the kernel size and $r$ the dilation rate, which equals one when nothing specified. Moreover, zero padding is applied to all layers to maintain spatial dimensions.

\subsection{Loss Function for Network Optimization}
The loss energy function proposed to train the model consists of two terms. The first term penalizes deviations between the reconstructed view and ground truth image:
\begin{equation}\label{eq:Ed}
    E_d = \|I_{p,q}-\hat{I}_{p,q}\|_1.
\end{equation}
To better preserve image textures, the second proposed term additionally imposes the output image to have similar spatial gradients to the ground truth:
\begin{equation}\label{eq:Eg}
    E_g = \|\nabla I_{p,q}-\nabla\hat{I}_{p,q}\|_1.
\end{equation}
Then, the proposed loss function writes as
\begin{equation}\label{eq:loss}
    E = E_d + \lambda_g E_g,
\end{equation}
where we experimentally set $\lambda_g = 0.5$. In Section \ref{sec:experiments} we evaluate different configurations for this training loss.

Another term we could have included is one that enforces consistency between different disparity maps, similar to~\cite{srinivasan2017learning}. However, disparity maps should not be equal at occluded regions and, since we do not know these occlusions beforehand, we do not impose any constraint.

\subsection{Training Details}
The model has been implemented using TensorFlow~\cite{abadi2016tensorflow}. We train the networks on Lytro light fields which have a spatial resolution of $540\times372$ and an angular one of $14\times14$, from which we select a centred $7\times7$ array of views. The four corner views of these $7\times7$ light fields are the inputs to our method. 

At each training iteration, we randomly select the angular coordinates at integer positions $p, q\in\mathbb{Z}\cap[0,6]$, excluding the corner views. The output is compared at each iteration to the ground truth view by means of the loss function presented in Equation~\eqref{eq:loss}. We randomly extract $192\times192$ patches from the training images to train the model. The network is optimized using the ADAM solver~\cite{kingma2014adam} with $\beta_1=0.9, \beta_2=0.999, \epsilon=1e-08$, a learning rate of $0.001$ and a batch size of 3. Weights are initialized randomly using the Xavier method~\cite{glorot2010understanding} and the softmax $\beta$ is initialized to 1. The method converges after $300$k iterations and it approximately takes 1 day and 20 hours on a GeForce GTX 1080 Ti GPU. At test time, it takes less than 2 seconds to synthesize a $540\times372$ image.

%%%%%%%%%%%%%%%%%%%%%%%%%%%% 
\begin{figure*}[t!]\centering
\begin{tabular}{@{}c@{\hskip 0.2em}c@{\hskip 0.2em}c@{\hskip 0.2em}c@{}}
\multicolumn{4}{c}{View $(1,1)$}
\\\noalign{\smallskip}
\multicolumn{2}{@{}c@{\hskip 0.1em}}{$I_{1,1}$} & 
\multicolumn{2}{@{}c@{}}{$\hat{I}_{1,1}$} 
\\
\multicolumn{2}{@{}c@{\hskip 0.1em}}{\includegraphics[width=0.15\linewidth]{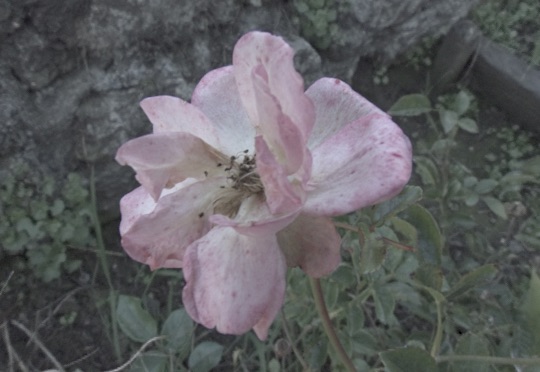}} &
\multicolumn{2}{@{}c@{}}{\includegraphics[width=0.15\linewidth]{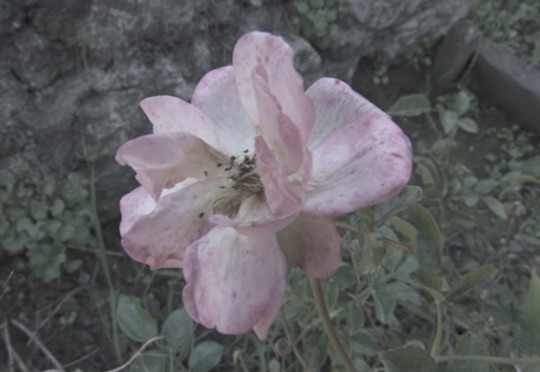}} 
\\
\includegraphics[width=0.073\linewidth,trim=95 42 295 180,clip=true]{figures/resultsLytro/IMG_7185_eslf/p1_q1/gt.jpg} &
\includegraphics[width=0.073\linewidth,trim=245 215 145 7,clip=true]{figures/resultsLytro/IMG_7185_eslf/p1_q1/gt.jpg} &
\includegraphics[width=0.073\linewidth,trim=95 42 295 180,clip=true]{figures/resultsLytro/IMG_7185_eslf/p1_q1/y.jpg} &
\includegraphics[width=0.073\linewidth,trim=245 215 145 7,clip=true]{figures/resultsLytro/IMG_7185_eslf/p1_q1/y.jpg} 
\\
\multicolumn{2}{@{}c@{\hskip 0.1em}}{$d_{0,0}$} &
\multicolumn{2}{@{}c@{}}{$d_{0,6}$} 	
\\
\multicolumn{2}{@{}c@{\hskip 0.1em}}{\includegraphics[width=0.15\linewidth]{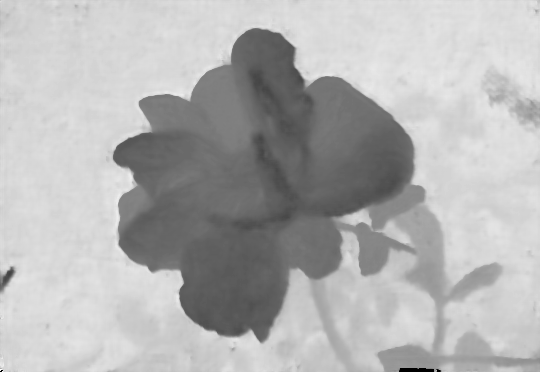}} &
\multicolumn{2}{@{}c@{\hskip 0.1em}}{\includegraphics[width=0.15\linewidth]{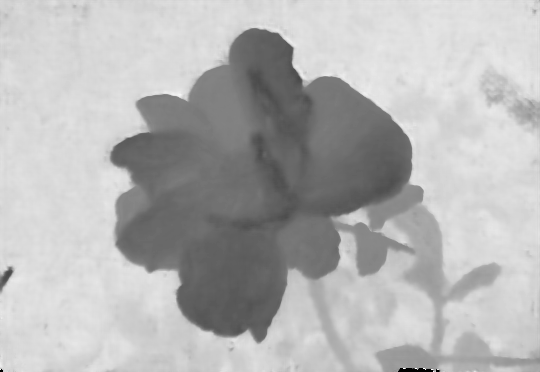}} 
\\
\includegraphics[width=0.073\linewidth,trim=95 42 295 180,clip=true]{figures/resultsLytro/IMG_7185_eslf/p1_q1/d00.jpg} &
\includegraphics[width=0.073\linewidth,trim=245 215 145 7,clip=true]{figures/resultsLytro/IMG_7185_eslf/p1_q1/d00.jpg} &
\includegraphics[width=0.073\linewidth,trim=95 42 295 180,clip=true]{figures/resultsLytro/IMG_7185_eslf/p1_q1/d01.jpg} &
\includegraphics[width=0.073\linewidth,trim=245 215 145 7,clip=true]{figures/resultsLytro/IMG_7185_eslf/p1_q1/d01.jpg} 
\\
\multicolumn{2}{@{}c@{\hskip 0.1em}}{$d_{6,0}$} &
\multicolumn{2}{@{}c@{}}{$d_{6,6}$} 	
\\
\multicolumn{2}{@{}c@{\hskip 0.1em}}{\includegraphics[width=0.15\linewidth]{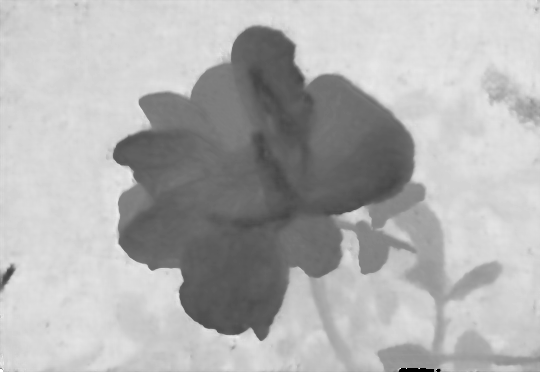}} &
\multicolumn{2}{@{}c@{\hskip 0.1em}}{\includegraphics[width=0.15\linewidth]{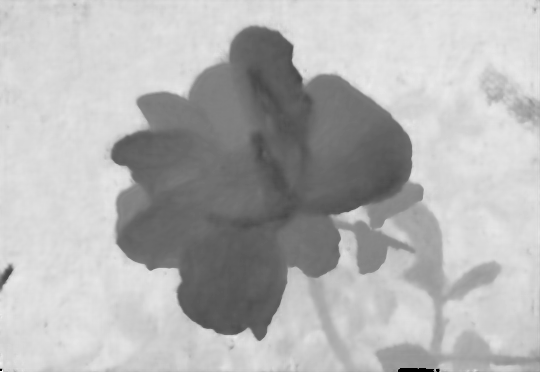}} 
\\
\includegraphics[width=0.073\linewidth,trim=95 42 295 180,clip=true]{figures/resultsLytro/IMG_7185_eslf/p1_q1/d10.jpg} &
\includegraphics[width=0.073\linewidth,trim=245 215 145 7,clip=true]{figures/resultsLytro/IMG_7185_eslf/p1_q1/d10.jpg} &
\includegraphics[width=0.073\linewidth,trim=95 42 295 180,clip=true]{figures/resultsLytro/IMG_7185_eslf/p1_q1/d11.jpg} &
\includegraphics[width=0.073\linewidth,trim=245 215 145 7,clip=true]{figures/resultsLytro/IMG_7185_eslf/p1_q1/d11.jpg} 
\\
\multicolumn{2}{@{}c@{\hskip 0.1em}}{$m_{0,0}$} &
\multicolumn{2}{@{}c@{}}{$m_{0,6}$} 	
\\
\multicolumn{2}{@{}c@{\hskip 0.1em}}{\includegraphics[width=0.15\linewidth]{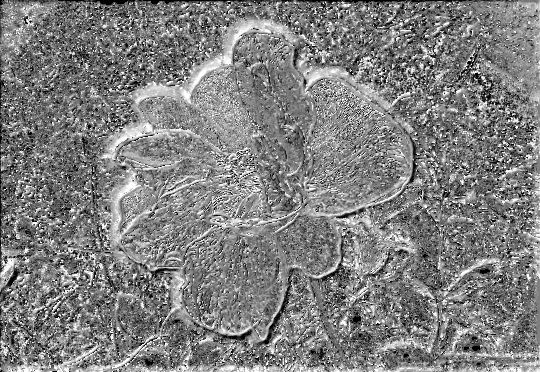}} &
\multicolumn{2}{@{}c@{\hskip 0.1em}}{\includegraphics[width=0.15\linewidth]{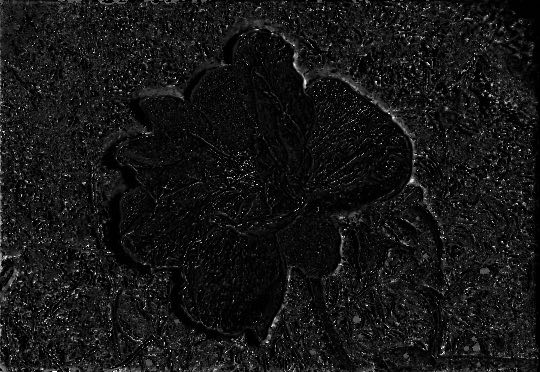}} 
\\
\includegraphics[width=0.073\linewidth,trim=95 42 295 180,clip=true]{figures/resultsLytro/IMG_7185_eslf/p1_q1/m00.jpg} &
\includegraphics[width=0.073\linewidth,trim=245 215 145 7,clip=true]{figures/resultsLytro/IMG_7185_eslf/p1_q1/m00.jpg} &
\includegraphics[width=0.073\linewidth,trim=95 42 295 180,clip=true]{figures/resultsLytro/IMG_7185_eslf/p1_q1/m01.jpg} &
\includegraphics[width=0.073\linewidth,trim=245 215 145 7,clip=true]{figures/resultsLytro/IMG_7185_eslf/p1_q1/m01.jpg} 
\\ 
\multicolumn{2}{@{}c@{\hskip 0.1em}}{$m_{6,0}$} &
\multicolumn{2}{@{}c@{}}{$m_{6,6}$} 	
\\
\multicolumn{2}{@{}c@{\hskip 0.1em}}{\includegraphics[width=0.15\linewidth]{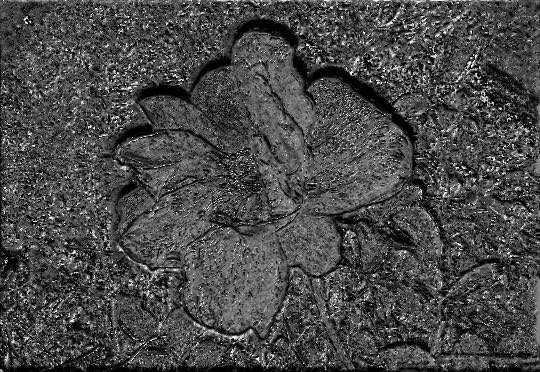}} &
\multicolumn{2}{@{}c@{\hskip 0.1em}}{\includegraphics[width=0.15\linewidth]{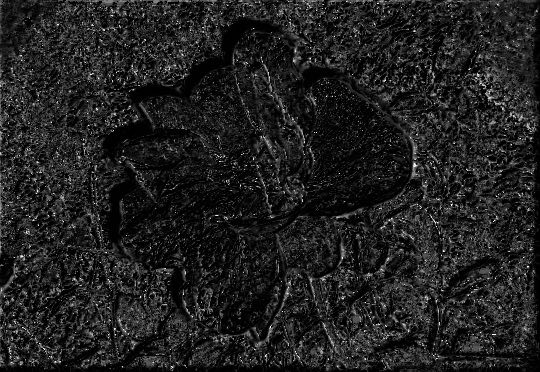}} 
\\
\includegraphics[width=0.073\linewidth,trim=95 42 295 180,clip=true]{figures/resultsLytro/IMG_7185_eslf/p1_q1/m10.jpg} &
\includegraphics[width=0.073\linewidth,trim=245 215 145 7,clip=true]{figures/resultsLytro/IMG_7185_eslf/p1_q1/m10.jpg} &
\includegraphics[width=0.073\linewidth,trim=95 42 295 180,clip=true]{figures/resultsLytro/IMG_7185_eslf/p1_q1/m11.jpg} &
\includegraphics[width=0.073\linewidth,trim=245 215 145 7,clip=true]{figures/resultsLytro/IMG_7185_eslf/p1_q1/m11.jpg} 
\end{tabular}
%%%
\hspace{3pt}
\begin{tabular}{@{}c@{\hskip 0.2em}c@{\hskip 0.2em}c@{\hskip 0.2em}c@{}}
\multicolumn{4}{c}{View $(3,3)$}
\\\noalign{\smallskip}
\multicolumn{2}{@{}c@{\hskip 0.1em}}{$I_{3,3}$} & 
\multicolumn{2}{@{}c@{\hskip 0.1em}}{$\hat{I}_{3,3}$} 
\\
\multicolumn{2}{@{}c@{\hskip 0.1em}}{\includegraphics[width=0.15\linewidth]{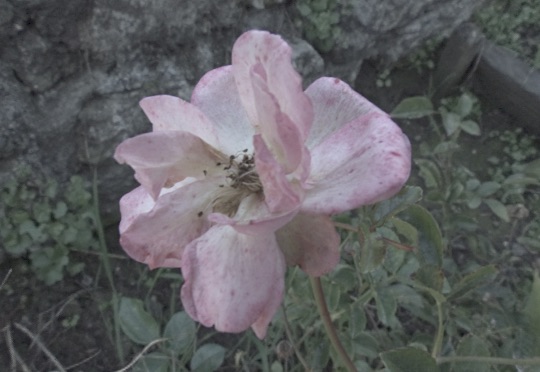}} &
\multicolumn{2}{@{}c@{\hskip 0.1em}}{\includegraphics[width=0.15\linewidth]{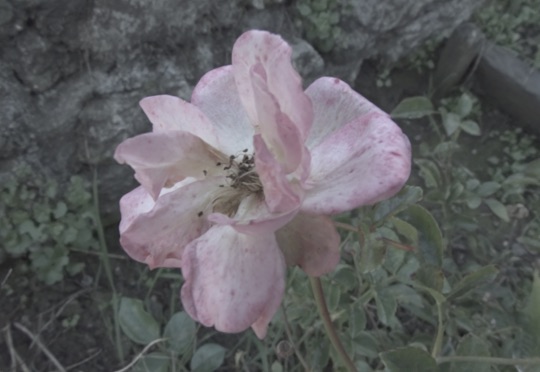}} 
\\
\includegraphics[width=0.073\linewidth,trim=95 42 295 180,clip=true]{figures/resultsLytro/IMG_7185_eslf/p3_q3/gt.jpg} &
\includegraphics[width=0.073\linewidth,trim=245 215 145 7,clip=true]{figures/resultsLytro/IMG_7185_eslf/p3_q3/gt.jpg} &
\includegraphics[width=0.073\linewidth,trim=95 42 295 180,clip=true]{figures/resultsLytro/IMG_7185_eslf/p3_q3/y.jpg} &
\includegraphics[width=0.073\linewidth,trim=245 215 145 7,clip=true]{figures/resultsLytro/IMG_7185_eslf/p3_q3/y.jpg} 
\\
\multicolumn{2}{@{}c@{\hskip 0.1em}}{$d_{0,0}$} &
\multicolumn{2}{@{}c@{}}{$d_{0,6}$} 	
\\
\multicolumn{2}{@{}c@{\hskip 0.1em}}{\includegraphics[width=0.15\linewidth]{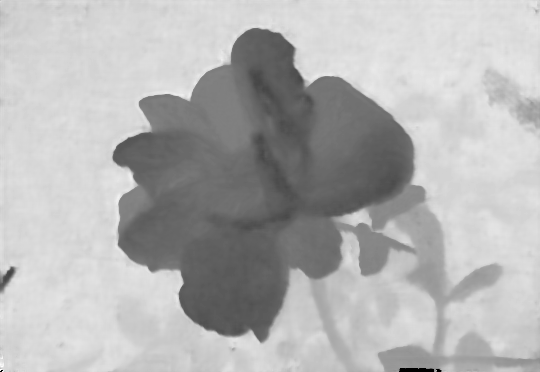}} &
\multicolumn{2}{@{}c@{\hskip 0.1em}}{\includegraphics[width=0.15\linewidth]{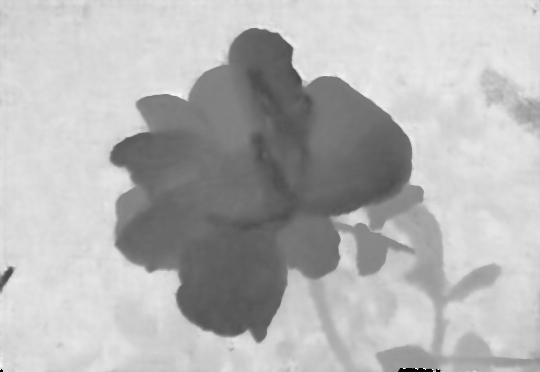}} 
\\
\includegraphics[width=0.073\linewidth,trim=95 42 295 180,clip=true]{figures/resultsLytro/IMG_7185_eslf/p3_q3/d00.jpg} &
\includegraphics[width=0.073\linewidth,trim=245 215 145 7,clip=true]{figures/resultsLytro/IMG_7185_eslf/p3_q3/d00.jpg} &
\includegraphics[width=0.073\linewidth,trim=95 42 295 180,clip=true]{figures/resultsLytro/IMG_7185_eslf/p3_q3/d01.jpg} &
\includegraphics[width=0.073\linewidth,trim=245 215 145 7,clip=true]{figures/resultsLytro/IMG_7185_eslf/p3_q3/d01.jpg} 
\\
\multicolumn{2}{@{}c@{\hskip 0.1em}}{$d_{6,0}$} &
\multicolumn{2}{@{}c@{}}{$d_{6,6}$} 	
\\
\multicolumn{2}{@{}c@{\hskip 0.1em}}{\includegraphics[width=0.15\linewidth]{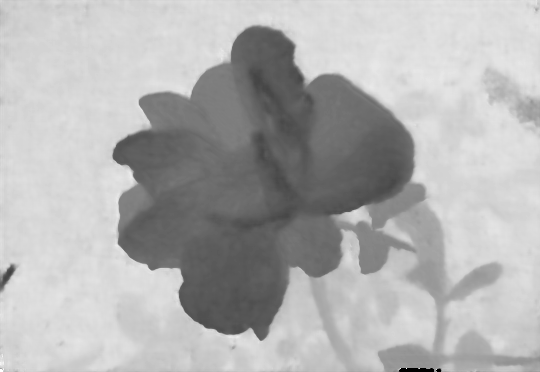}} &
\multicolumn{2}{@{}c@{\hskip 0.1em}}{\includegraphics[width=0.15\linewidth]{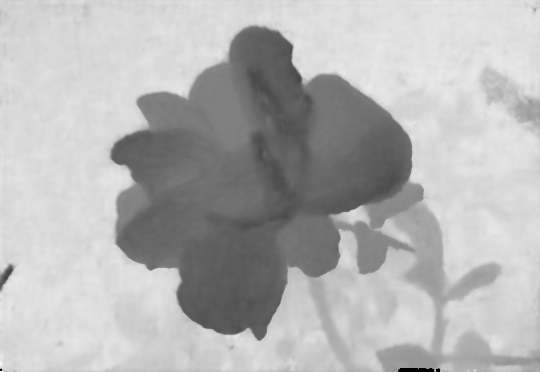}} 
\\
\includegraphics[width=0.073\linewidth,trim=95 42 295 180,clip=true]{figures/resultsLytro/IMG_7185_eslf/p3_q3/d10.jpg} &
\includegraphics[width=0.073\linewidth,trim=245 215 145 7,clip=true]{figures/resultsLytro/IMG_7185_eslf/p3_q3/d10.jpg} &
\includegraphics[width=0.073\linewidth,trim=95 42 295 180,clip=true]{figures/resultsLytro/IMG_7185_eslf/p3_q3/d11.jpg} &
\includegraphics[width=0.073\linewidth,trim=245 215 145 7,clip=true]{figures/resultsLytro/IMG_7185_eslf/p3_q3/d11.jpg} 
\\
\multicolumn{2}{@{}c@{\hskip 0.1em}}{$m_{0,0}$} &
\multicolumn{2}{@{}c@{}}{$m_{0,6}$} 	
\\
\multicolumn{2}{@{}c@{\hskip 0.1em}}{\includegraphics[width=0.15\linewidth]{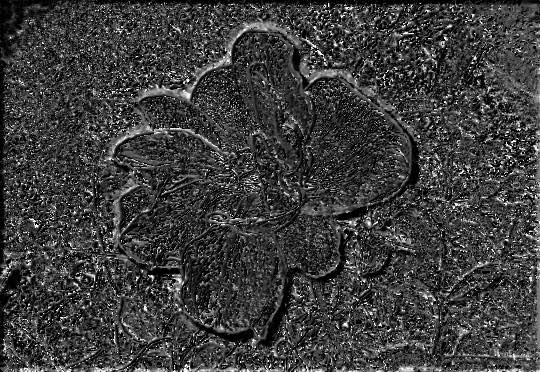}} &
\multicolumn{2}{@{}c@{\hskip 0.1em}}{\includegraphics[width=0.15\linewidth]{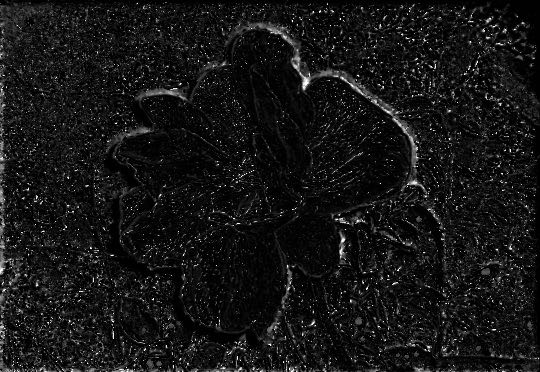}} 
\\
\includegraphics[width=0.073\linewidth,trim=95 42 295 180,clip=true]{figures/resultsLytro/IMG_7185_eslf/p3_q3/m00.jpg} &
\includegraphics[width=0.073\linewidth,trim=245 215 145 7,clip=true]{figures/resultsLytro/IMG_7185_eslf/p3_q3/m00.jpg} &
\includegraphics[width=0.073\linewidth,trim=95 42 295 180,clip=true]{figures/resultsLytro/IMG_7185_eslf/p3_q3/m01.jpg} &
\includegraphics[width=0.073\linewidth,trim=245 215 145 7,clip=true]{figures/resultsLytro/IMG_7185_eslf/p3_q3/m01.jpg} 
\\
\multicolumn{2}{@{}c@{\hskip 0.1em}}{$m_{6,0}$} &
\multicolumn{2}{@{}c@{}}{$m_{6,6}$} 	
\\
\multicolumn{2}{@{}c@{\hskip 0.1em}}{\includegraphics[width=0.15\linewidth]{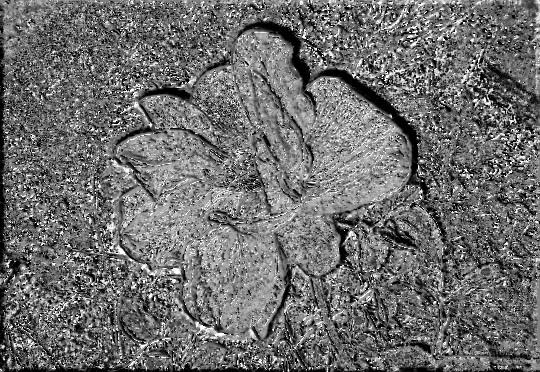}} &
\multicolumn{2}{@{}c@{\hskip 0.1em}}{\includegraphics[width=0.15\linewidth]{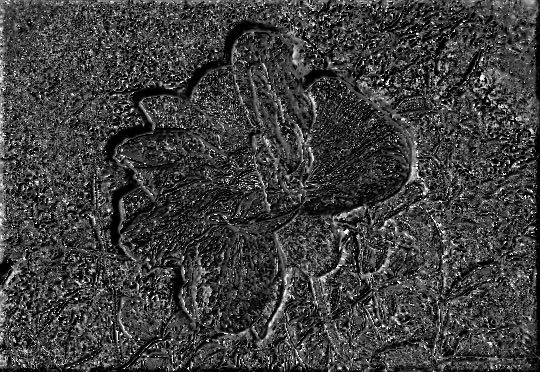}} 
\\
\includegraphics[width=0.073\linewidth,trim=95 42 295 180,clip=true]{figures/resultsLytro/IMG_7185_eslf/p3_q3/m10.jpg} &
\includegraphics[width=0.073\linewidth,trim=245 215 145 7,clip=true]{figures/resultsLytro/IMG_7185_eslf/p3_q3/m10.jpg} &
\includegraphics[width=0.073\linewidth,trim=95 42 295 180,clip=true]{figures/resultsLytro/IMG_7185_eslf/p3_q3/m11.jpg} &
\includegraphics[width=0.073\linewidth,trim=245 215 145 7,clip=true]{figures/resultsLytro/IMG_7185_eslf/p3_q3/m11.jpg} 
\end{tabular}
%%%
\hspace{3pt}
\begin{tabular}{@{}c@{\hskip 0.2em}c@{\hskip 0.2em}c@{\hskip 0.2em}c@{}}
\multicolumn{4}{c}{View $(5,5)$}
\\\noalign{\smallskip}
\multicolumn{2}{@{}c@{\hskip 0.1em}}{$I_{5,5}$} & 
\multicolumn{2}{@{}c@{\hskip 0.1em}}{$\hat{I}_{5,5}$} 
\\
\multicolumn{2}{@{}c@{\hskip 0.1em}}{\includegraphics[width=0.15\linewidth]{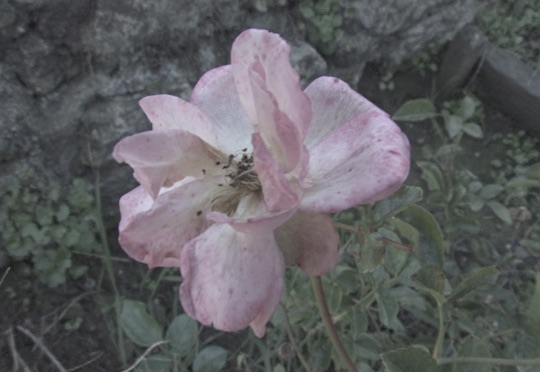}} &
\multicolumn{2}{@{}c@{\hskip 0.1em}}{\includegraphics[width=0.15\linewidth]{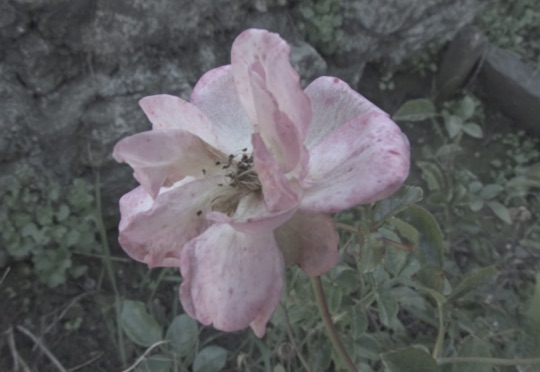}} 
\\
\includegraphics[width=0.073\linewidth,trim=95 42 295 180,clip=true]{figures/resultsLytro/IMG_7185_eslf/p5_q5/gt.jpg} &
\includegraphics[width=0.073\linewidth,trim=245 215 145 7,clip=true]{figures/resultsLytro/IMG_7185_eslf/p5_q5/gt.jpg} &
\includegraphics[width=0.073\linewidth,trim=95 42 295 180,clip=true]{figures/resultsLytro/IMG_7185_eslf/p5_q5/y.jpg} &
\includegraphics[width=0.073\linewidth,trim=245 215 145 7,clip=true]{figures/resultsLytro/IMG_7185_eslf/p5_q5/y.jpg} 
\\
\multicolumn{2}{@{}c@{\hskip 0.1em}}{$d_{0,0}$} &
\multicolumn{2}{@{}c@{}}{$d_{0,6}$} 	
\\
\multicolumn{2}{@{}c@{\hskip 0.1em}}{\includegraphics[width=0.15\linewidth]{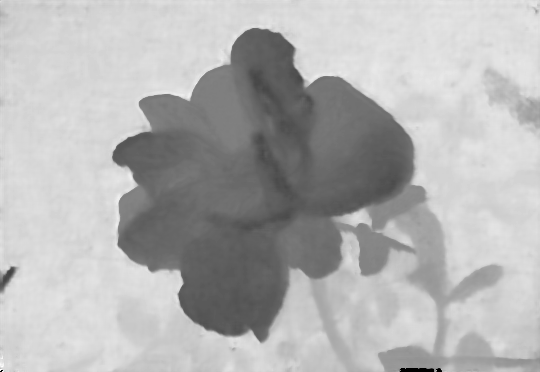}} &
\multicolumn{2}{@{}c@{\hskip 0.1em}}{\includegraphics[width=0.15\linewidth]{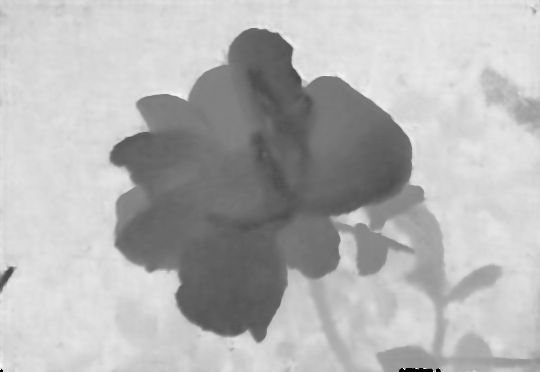}} 
\\
\includegraphics[width=0.073\linewidth,trim=95 42 295 180,clip=true]{figures/resultsLytro/IMG_7185_eslf/p5_q5/d00.jpg} &
\includegraphics[width=0.073\linewidth,trim=245 215 145 7,clip=true]{figures/resultsLytro/IMG_7185_eslf/p5_q5/d00.jpg} &
\includegraphics[width=0.073\linewidth,trim=95 42 295 180,clip=true]{figures/resultsLytro/IMG_7185_eslf/p5_q5/d01.jpg} &
\includegraphics[width=0.073\linewidth,trim=245 215 145 7,clip=true]{figures/resultsLytro/IMG_7185_eslf/p5_q5/d01.jpg} 
\\
\multicolumn{2}{@{}c@{\hskip 0.1em}}{$d_{6,0}$} &
\multicolumn{2}{@{}c@{}}{$d_{6,6}$} 	
\\
\multicolumn{2}{@{}c@{\hskip 0.1em}}{\includegraphics[width=0.15\linewidth]{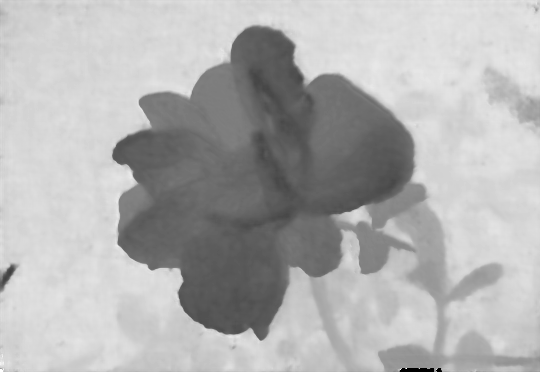}} &
\multicolumn{2}{@{}c@{\hskip 0.1em}}{\includegraphics[width=0.15\linewidth]{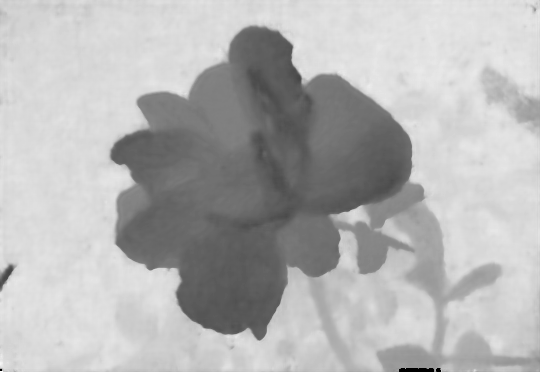}} 
\\
\includegraphics[width=0.073\linewidth,trim=95 42 295 180,clip=true]{figures/resultsLytro/IMG_7185_eslf/p5_q5/d10.jpg} &
\includegraphics[width=0.073\linewidth,trim=245 215 145 7,clip=true]{figures/resultsLytro/IMG_7185_eslf/p5_q5/d10.jpg} &
\includegraphics[width=0.073\linewidth,trim=95 42 295 180,clip=true]{figures/resultsLytro/IMG_7185_eslf/p5_q5/d11.jpg} &
\includegraphics[width=0.073\linewidth,trim=245 215 145 7,clip=true]{figures/resultsLytro/IMG_7185_eslf/p5_q5/d11.jpg} 
\\
\multicolumn{2}{@{}c@{\hskip 0.1em}}{$m_{0,0}$} &
\multicolumn{2}{@{}c@{}}{$m_{0,6}$} 	
\\
\multicolumn{2}{@{}c@{\hskip 0.1em}}{\includegraphics[width=0.15\linewidth]{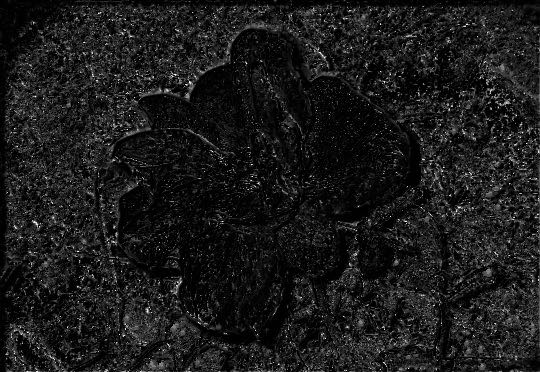}} &
\multicolumn{2}{@{}c@{\hskip 0.1em}}{\includegraphics[width=0.15\linewidth]{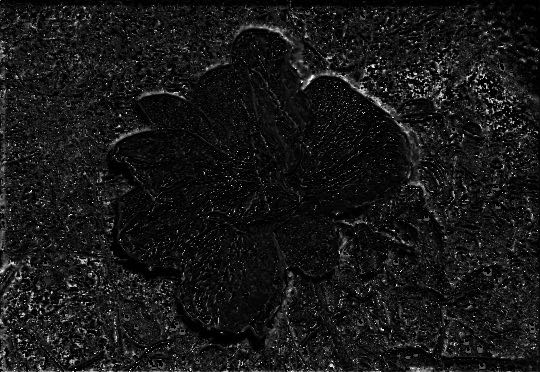}} 
\\
\includegraphics[width=0.073\linewidth,trim=95 42 295 180,clip=true]{figures/resultsLytro/IMG_7185_eslf/p5_q5/m00.jpg} &
\includegraphics[width=0.073\linewidth,trim=245 215 145 7,clip=true]{figures/resultsLytro/IMG_7185_eslf/p5_q5/m00.jpg} &
\includegraphics[width=0.073\linewidth,trim=95 42 295 180,clip=true]{figures/resultsLytro/IMG_7185_eslf/p5_q5/m01.jpg} &
\includegraphics[width=0.073\linewidth,trim=245 215 145 7,clip=true]{figures/resultsLytro/IMG_7185_eslf/p5_q5/m01.jpg} 
\\
\multicolumn{2}{@{}c@{\hskip 0.1em}}{$m_{6,0}$} &
\multicolumn{2}{@{}c@{}}{$m_{6,6}$} 	
\\
\multicolumn{2}{@{}c@{\hskip 0.1em}}{\includegraphics[width=0.15\linewidth]{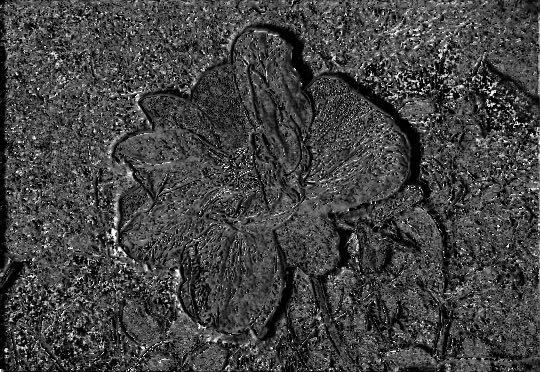}} &
\multicolumn{2}{@{}c@{\hskip 0.1em}}{\includegraphics[width=0.15\linewidth]{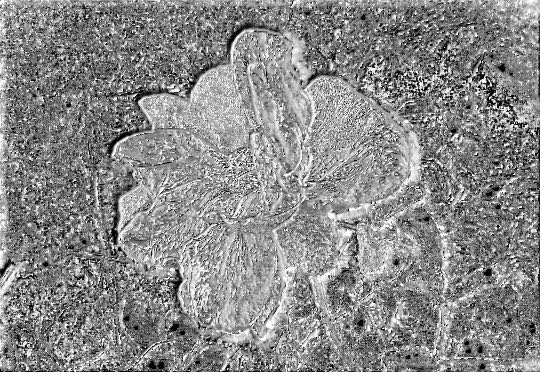}} 
\\
\includegraphics[width=0.073\linewidth,trim=95 42 295 180,clip=true]{figures/resultsLytro/IMG_7185_eslf/p5_q5/m10.jpg} &
\includegraphics[width=0.073\linewidth,trim=245 215 145 7,clip=true]{figures/resultsLytro/IMG_7185_eslf/p5_q5/m10.jpg} &
\includegraphics[width=0.073\linewidth,trim=95 42 295 180,clip=true]{figures/resultsLytro/IMG_7185_eslf/p5_q5/m11.jpg} &
\includegraphics[width=0.073\linewidth,trim=245 215 145 7,clip=true]{figures/resultsLytro/IMG_7185_eslf/p5_q5/m11.jpg} 
\end{tabular}
\caption{Visual results for three different input angular positions for the novel view. We display the ground truth and estimated view (top), the predicted disparity maps for each corner image (middle) and the four selection masks (bottom). We encourage the reader to look at the electronic version of this paper to better see the details.} %Testing light field corresponds to {\it IMG\_7185\_eslf} from the {\it Flowers} dataset. 
\label{fig:proposed}
\end{figure*}
%%%%%%%%%%%%%%%%%%%%%%%%%%%% 

%%%%%%%%%%%%%%%%%%%%%%%%%%%%%%%%%%%%%%%%%%
%%															  %%
%%						EXPERIMENTS				        	           %%
%%														 	  %%
%%%%%%%%%%%%%%%%%%%%%%%%%%%%%%%%%%%%%%%%%%
\section{Experiments}\label{sec:experiments}
In this section, we evaluate the performance of the proposed method. First, we assess the different components included in our approach by means of several experiments. Then, we compare the obtained results against state-of-the-art methods for light field view synthesis. Finally, the method is adapted and tested on light fields acquired with an array of cameras.

The quantitative evaluation reported during this section is in terms of the mean absolute error, which is multiplied by 100 for images in the range $[0,1]$ (MAE), the peak signal-to-noise ratio (PSNR) and the structural similarity index measure (SSIM)~\cite{wang2004image}. Unless otherwise stated, the reported metrics are averaged over all possible in-between viewpoints at integer positions, excluding the input corner ones, and over the whole test set under consideration. Besides, when nothing specified, all presented visual results correspond to the center view, which has angular coordinates $(3,3)$. Finally, for the sake of simplicity, in some cases we just display one disparity map as the disparity estimated by the proposed method, which actually corresponds to $d_{0,0}$.

\subsection{Datasets}
We used two different datasets for training and testing the proposed method. On the one hand, the dataset from Srinivasan et al.~\cite{srinivasan2017learning}, which consists of 3343 images of flowers captured with a Lytro Illum camera. We randomly divided it into 3243 images for training and 100 for testing the model. On the other hand, the dataset from Kalantari et al.~\cite{kalantari2016learning}. It contains 100 light fields for training and 30 for testing. They are mostly outdoor images from diverse scenarios captured with the Lytro Illum. When reading the images, we apply a gamma correction with $\gamma=0.4$ to both datasets. We denote as {\it Flowers} the dataset from Srinivasan et al.~\cite{srinivasan2017learning} and as {\it Diverse} the one from Kalantari et al.~\cite{kalantari2016learning}. For the experiments in this section, when nothing specified, the used dataset for training is {\it Flowers}.

\subsection{Visual Results} 
Figure~\ref{fig:proposed} visually illustrates the performance of the proposed model on one example of the {\it Flowers} test set and for three different angular coordinates for the novel view. As it can be seen in the figure, disparity maps are sharp at all depth discontinuities but they are more blurred at occlusions. At occluded regions, the warped views will be inaccurate. However, with the selection network we are able to discard occluded pixels. Occluded parts are equal to zero in the selection masks and more weight is given to the areas that are visible in only one view. Also, we can appreciate how the selection network has a preference on choosing the warped view whose angular position is closest to the novel one. This occurs because Lytro light fields present changes in color between views and the closer the viewpoints are, the more similar color the images have.

%%%%%%%%%%%%%%%%%%%%%%%%%%%%
\begin{table}[t!]\centering
\caption{Analysis of different terms in the loss function.}
\resizebox{\columnwidth}{!}{
\begin{tabular}{|c|c|c|c|c|c|c|c|c|}  \cline{4-9}
\multicolumn{1}{c}{}&\multicolumn{1}{c}{}&\multicolumn{1}{c}{}& \multicolumn{3}{|c|}{\it Flowers} & \multicolumn{3}{c|}{\it Diverse}\\ \cline{4-9} \noalign{\smallskip}\hline 

$E_d$		& $E_g $  		& $E_w$       	&MAE	& PSNR  		& SSIM &MAE     & PSNR  & SSIM   \\ \hline \hline 
$\checkmark$ &    			& 			&0.887	&38.07	&0.9770	&0.820	&37.82	&0.9834\\ \hline
$\checkmark$ & $\checkmark$ &  			&\bf0.879	&\bf38.28	&\bf0.9778	&\bf0.799	&\bf38.12	&\bf0.9848\\ \hline
$\checkmark$ & $\checkmark$ &$\checkmark$ &0.934	&37.74	&0.9757	&0.820	&37.87	&0.9846\\	\hline
\end{tabular}}
\label{tab:loss}
\end{table}
%%%%%%%%%%%%%%%%%%%%%%%%%%%%

\subsection{Analysis of the Loss Function}
Table \ref{tab:loss} reports evaluation metrics for three different configurations of the training loss. First, the model trained with the only use of the reconstruction error term $E_d$~\eqref{eq:Ed}. Second, using the proposed $E_d$ and the gradients difference term $E_g$~\eqref{eq:Eg}. Third, apart from $E_d$ and $E_g$, we additionally include a term $E_w$ that enforces each disparity estimation to be consistent with the warped views. That is, 
\begin{equation}\label{eq:loss_w}
    E_w = \frac{1}{4}\sum_{i,j\in\{0,N\}} \|I_{p,q}-I^w_{i,j}\|_1.
\end{equation}
As it is reported in the table, the proposed loss function combining just the reconstruction error and gradients differences outperforms the other settings in both test datasets.

%%%%%%%%%%%%%%%%%%%%%%%%%%%% 
\begin{table}[t!]\centering
\caption{Comparison against one single network, the use of just one disparity map and without the use of the features network.}
\resizebox{\columnwidth}{!}{
\begin{tabular}{|@{\hskip0.07in}c@{\hskip0.07in}|@{\hskip0.07in}c@{\hskip0.07in}|@{\hskip0.07in}c@{\hskip0.07in}|c@{\hskip0.07in}|@{\hskip0.07in}c@{\hskip0.07in}|@{\hskip0.07in}c@{\hskip0.07in}|@{\hskip0.07in}c@{\hskip0.07in}|@{\hskip0.07in}c@{\hskip0.07in}|@{\hskip0.07in}c@{\hskip0.07in}|}\cline{4-9}
\multicolumn{2}{c}{} & & \multicolumn{3}{@{\hskip0.07in}@{\hskip0.07in}c@{\hskip0.07in}|@{\hskip0.07in}}{\it Flowers} & \multicolumn{3}{c|}{\it Diverse} \\ \cline{4-9} \noalign{\smallskip}\hline

 Method & Param.   &  Time & MAE   & PSNR  & SSIM   & MAE   & PSNR  & SSIM  \\ \hline\hline
1 CNN      & 1.66 M  		&1.40 s	& 10.69	&24.54 & 0.9488 & 11.62 & 23.86 &  0.9493 \\ \hline
1 disp.      	& 1.27 M       	&1.91 s	&0.931 &37.76  &0.9750 &1.030 &36.03 &0.9732 \\ \hline
w/o $f_s$	& 1.27 M       	&1.82 s	&0.931 &37.81  &0.9757 &1.080 &35.66 &0.9703 \\ \hline
Proposed	& 1.27 M     	&1.89 s	&\bf0.879	&\bf38.28	& \bf0.9778 &\bf0.799 &\bf38.12 &\bf0.9848  \\ \hline
\end{tabular}}
\label{tab:othermodels}
\end{table}
%%%%%%%%%%%%%%%%%%%%%%%%%%%% 

%%%%%%%%%%%%%%%%%%%%%%%%%%%% 
\begin{figure}[t!]\centering
\begin{tabular}{@{}c@{\hskip 0.2em}c@{\hskip 0.2em}c@{\hskip 0.2em}c@{\hskip 0.2em}c@{\hskip 0.2em}c@{}}
\multicolumn{2}{@{}c@{\hskip 0.1em}}{Ground truth} & 
\multicolumn{2}{@{\hskip 0.1em}c@{\hskip 0.1em}}{Single CNN} & 
\multicolumn{2}{@{\hskip 0.1em}c@{}}{Proposed} 
\\
\includegraphics[width=0.16\linewidth,trim=130 162 340 140,clip=true]{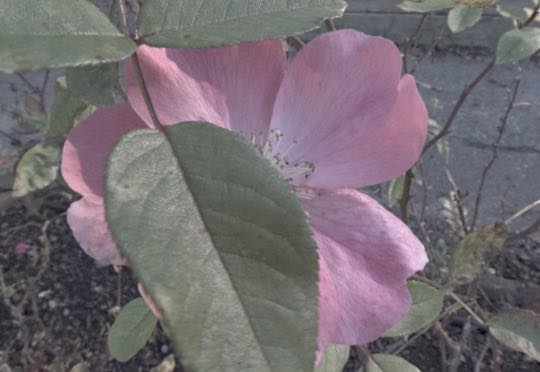} &
\includegraphics[width=0.16\linewidth,trim=350 162 120 140,clip=true]{figures/1cnn/7383_gt.jpg} &
\includegraphics[width=0.16\linewidth,trim=130 162 340 140,clip=true]{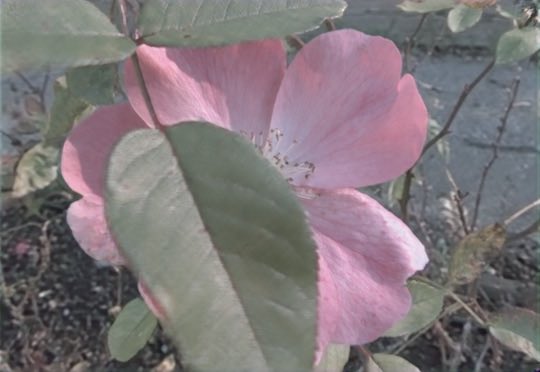} &
\includegraphics[width=0.16\linewidth,trim=350 162 120 140,clip=true]{figures/1cnn/7383_1cnn.jpg} &
\includegraphics[width=0.16\linewidth,trim=130 162 340 140,clip=true]{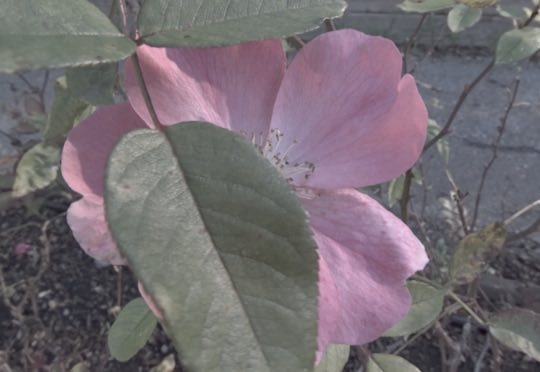} &
\includegraphics[width=0.16\linewidth,trim=350 162 120 140,clip=true]{figures/1cnn/7383_ours.jpg} 
\\
\includegraphics[width=0.16\linewidth,trim=130 162 340 140,clip=true]{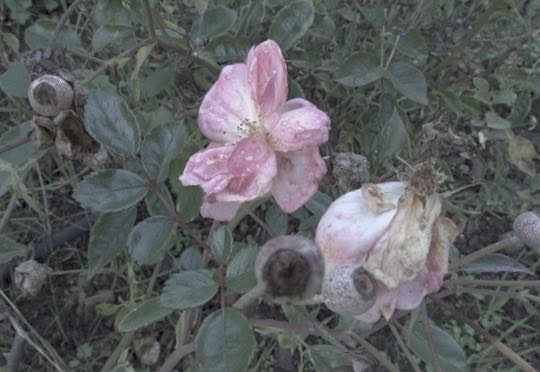} &
\includegraphics[width=0.16\linewidth,trim=270 102 200 200,clip=true]{figures/1cnn/7005_gt.jpg} &
\includegraphics[width=0.16\linewidth,trim=130 162 340 140,clip=true]{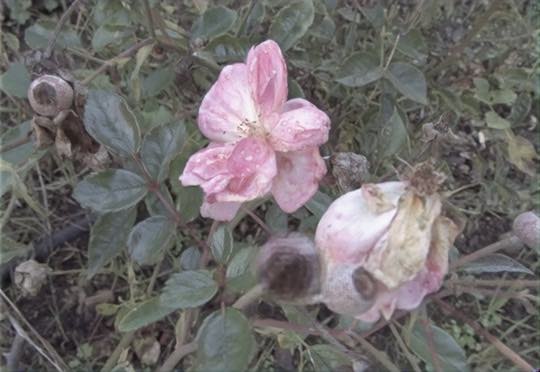} &
\includegraphics[width=0.16\linewidth,trim=270 102 200 200,clip=true]{figures/1cnn/7005_1cnn.jpg} &
\includegraphics[width=0.16\linewidth,trim=130 162 340 140,clip=true]{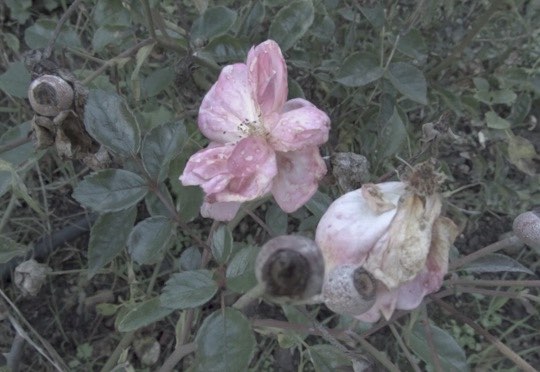} &
\includegraphics[width=0.16\linewidth,trim=270 102 200 200,clip=true]{figures/1cnn/7005_ours.jpg} 
\end{tabular}
\caption{Comparison against one single network. The single network has poorer textures and is unable to correctly model the geometry of the scene.}
\label{fig:1cnn} %Images correspond to {\it IMG\_7383\_eslf} and {\it IMG\_7005\_eslf} from the {\it Flowers} dataset.
\end{figure}
%%%%%%%%%%%%%%%%%%%%%%%%%%%% 

\subsection{Comparison with One Single CNN}
We compare the proposed approach against using one single CNN to model the view synthesis problem. The implemented single-CNN model consists of a fully-convolutional network of 22 layers with kernel sizes of $3\times3$. Also, as in our disparity network, we use dilated convolutions from the fourth to the seventh layers at rates $2, 4, 8$ and $16$, respectively. This results in a network with $1.66$ millions of parameters. % and a receptive field of $97$ pixels. %Thus, we believe it is comparable to the proposed approach.
The two images $P$ and $Q$ containing the angular coordinates and the four corner views are concatenated along the channels dimension and are the input to the network. The output is the color novel view at the indicated position. 
%The loss function considered for this single network is the same as $E=E_d+\lambda_g E_g$, with $E_d$ and $E_g$ being the ones from Equations~\eqref{eq:Ed} and \eqref{eq:Eg}, respectively, and $\lambda_g=0.5$. 
We have trained this model using the {\it Flowers} training set.

In Table~\ref{tab:othermodels} we quantitatively compare both models. The single CNN takes in average half a second less than the proposed approach but the performance is significantly worse. As seen in Figure~\ref{fig:1cnn}, the single CNN reconstruction results are blurry and the network is unable to correctly model the geometry of the scene.

%%%%%%%%%%%%%%%%%%%%%%%%%%%% 
\begin{figure}[t!]\centering
\begin{tabular}{@{}c@{\hskip 0.1em}c@{\hskip 0.1em}c@{\hskip 0.1em}c@{\hskip 0.1em}c@{\hskip 0.1em}c@{\hskip 0.1em}c@{\hskip 0.1em}c@{}}
\multicolumn{2}{@{}c@{\hskip 0.1em}}{Ground truth} & 
\multicolumn{2}{@{}c@{\hskip 0.1em}}{1 disparity}&
\multicolumn{2}{@{}c@{\hskip 0.1em}}{$d_{0,0}$} &
\multicolumn{2}{@{}c@{}}{$d_{0,6}$} 	
\\
\multicolumn{2}{@{}c@{\hskip 0.1em}}{\includegraphics[width=0.249\linewidth]{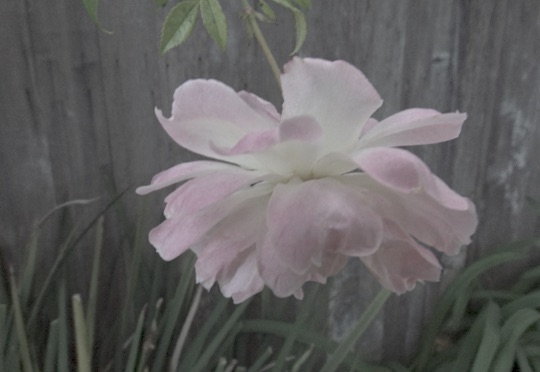}} &
\multicolumn{2}{@{}c@{\hskip 0.1em}}{\includegraphics[width=0.249\linewidth]{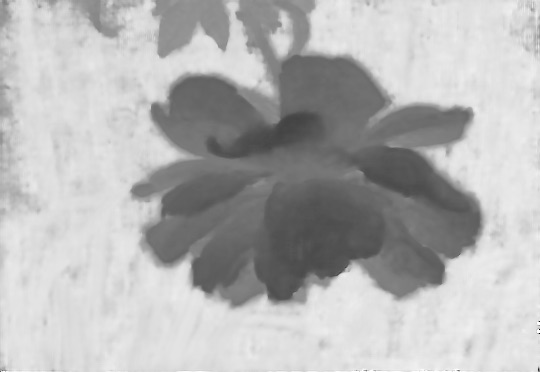}} &
\multicolumn{2}{@{}c@{\hskip 0.1em}}{\includegraphics[width=0.249\linewidth]{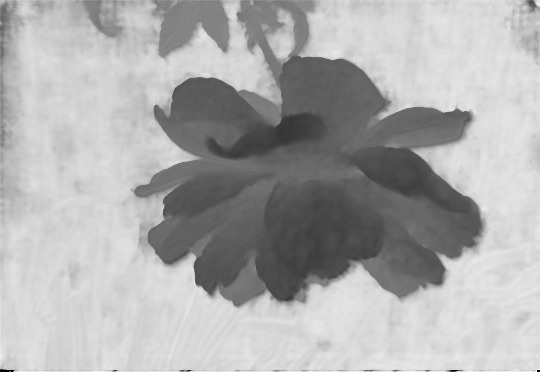}} &
\multicolumn{2}{@{}c@{}}{\includegraphics[width=0.249\linewidth]{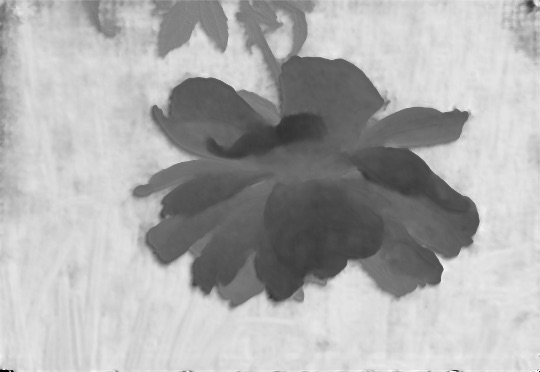}} 
\\
\includegraphics[width=0.12\linewidth,trim=135 92 335 210,clip=true]{figures/1disp/5102_gt.jpg} &
\includegraphics[width=0.12\linewidth,trim=385 65 85 237,clip=true]{figures/1disp/5102_gt.jpg} &
\includegraphics[width=0.12\linewidth,trim=135 92 335 210,clip=true]{figures/1disp/5102_1disp_d.jpg} &
\includegraphics[width=0.12\linewidth,trim=385 65 85 237,clip=true]{figures/1disp/5102_1disp_d.jpg} &
\includegraphics[width=0.12\linewidth,trim=135 92 335 210,clip=true]{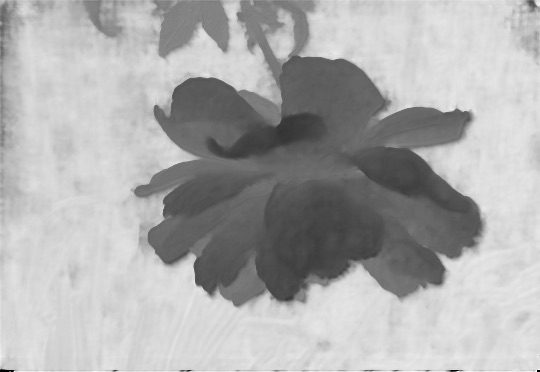} &
\includegraphics[width=0.12\linewidth,trim=385 65 85 237,clip=true]{figures/1disp/d00.jpg} &
\includegraphics[width=0.12\linewidth,trim=135 92 335 210,clip=true]{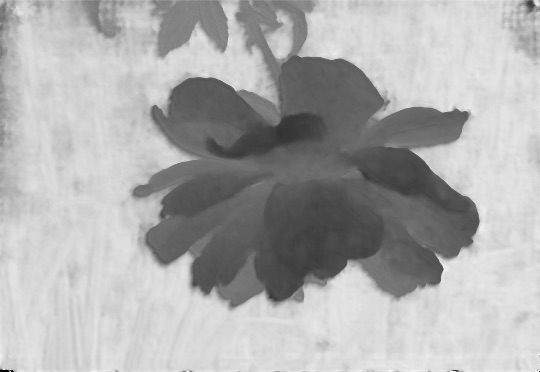} &
\includegraphics[width=0.12\linewidth,trim=385 65 85 237,clip=true]{figures/1disp/d01.jpg} 
\\
\multicolumn{2}{c}{Error 1 disp.} &
\multicolumn{2}{c}{Error 4 disp.} &
\multicolumn{2}{c}{$d_{6,0}$} &
\multicolumn{2}{c}{$d_{6,6}$} 
\\
\multicolumn{2}{@{}c@{\hskip 0.1em}}{\includegraphics[width=0.249\linewidth]{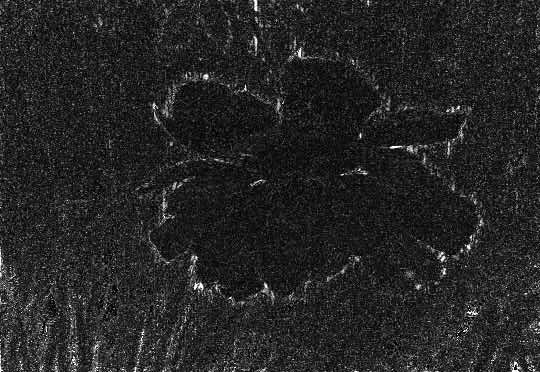}} &
\multicolumn{2}{@{}c@{\hskip 0.1em}}{\includegraphics[width=0.249\linewidth]{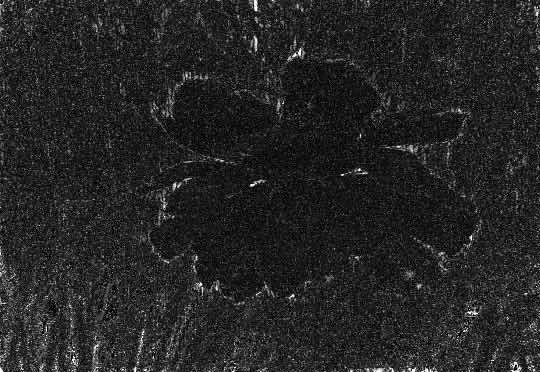}}&
\multicolumn{2}{@{}c@{\hskip 0.1em}}{\includegraphics[width=0.249\linewidth]{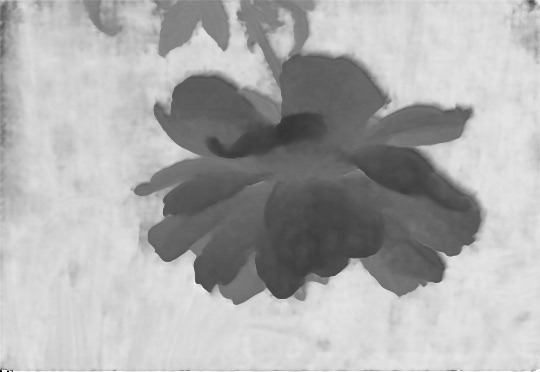}} &         
\multicolumn{2}{@{}c@{}}{\includegraphics[width=0.249\linewidth]{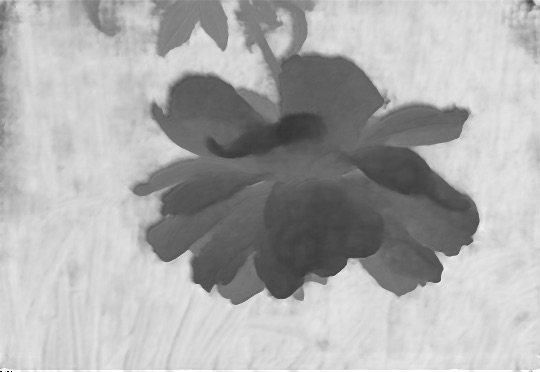}} 
\\
\includegraphics[width=0.12\linewidth,trim=135 92 335 210,clip=true]{figures/1disp/5102_1disp_error.jpg} &
\includegraphics[width=0.12\linewidth,trim=385 65 85 237,clip=true]{figures/1disp/5102_1disp_error.jpg} &
\includegraphics[width=0.12\linewidth,trim=135 92 335 210,clip=true]{figures/1disp/5102_ours_error.jpg} &
\includegraphics[width=0.12\linewidth,trim=385 65 85 237,clip=true]{figures/1disp/5102_ours_error.jpg} &
\includegraphics[width=0.12\linewidth,trim=135 92 335 210,clip=true]{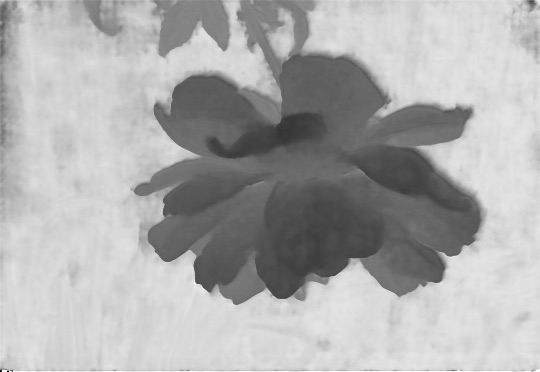} &
\includegraphics[width=0.12\linewidth,trim=385 65 85 237,clip=true]{figures/1disp/d10.jpg} &
\includegraphics[width=0.12\linewidth,trim=135 92 335 210,clip=true]{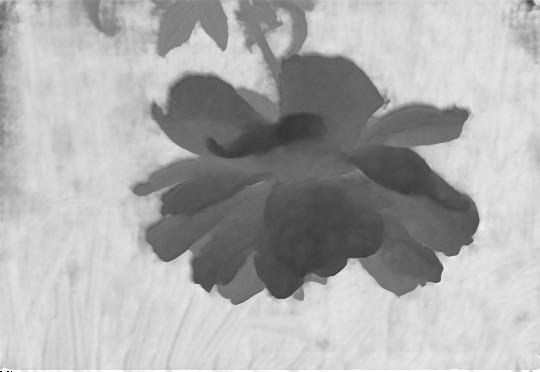} &
\includegraphics[width=0.12\linewidth,trim=385 65 85 237,clip=true]{figures/1disp/d11.jpg} 
\end{tabular}
\caption{Effect of using one common disparity map or the proposed multiple disparity maps. Reconstruction errors are clipped into the range $[0,0.04]$ for images in $[0,1]$. We encourage the reader to look at the electronic version of this paper to better see the details.}
\label{fig:1vs4disps}
\end{figure}
%%%%%%%%%%%%%%%%%%%%%%%%%%%% 

\subsection{Advantage of Using Four Disparity Maps}
Next, we show the importance of considering four different disparity estimations compared to the use of one common disparity map for the four views, as it is done in~\cite{kalantari2016learning}. In Table~\ref{tab:othermodels} we quantitatively compare these two strategies. The use of multiple disparity estimations improves the estimation. In the {\it Flowers} dataset we obtain an average PSNR of $38.28$, while with one common disparity it decreases to $37.76$. On the other hand, in the {\it Diverse} test set we maintain a high PSNR of $38.12$ with the proposed method, opposed to the $36.03$ yielded by the use of one single disparity.

Figure~\ref{fig:1vs4disps} illustrates this comparison. The use of a common disparity leads to inaccuracies in the estimation that are located at the union of the occluded parts of the four images. That is, for instance, next to the boundaries of the flower. However, if we look to the case of having four disparities, we can see how the areas of the views corresponding to non-occluded regions are sharp and accurate, while the occluded parts present more difficulties. The effect in the final result is reflected in the error images, where the errors at occlusions are significantly smaller in the case of using four disparities.

%%%%%%%%%%%%%%%%%%%%%%%%%%%% 
\begin{figure}[t!]\centering
\begin{tabular}{@{}c@{\hskip0.2em}c@{\hskip0.2em}c@{\hskip0.2em}c@{\hskip0.2em}c@{}}
\multicolumn{2}{c}{Without the features CNN} & \multicolumn{2}{c}{With the features CNN}
\\
\includegraphics[width=0.23\linewidth,trim=50 112 380 150,clip=true]{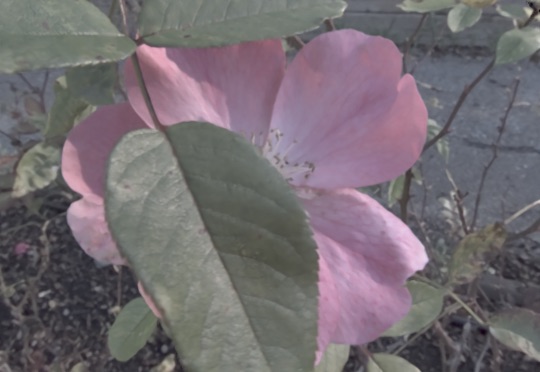} &
\includegraphics[width=0.23\linewidth,trim=240 132 190 130,clip=true]{nofcnn/7383_nof_y.jpg} &
\includegraphics[width=0.23\linewidth,trim=50 112 380 150,clip=true]{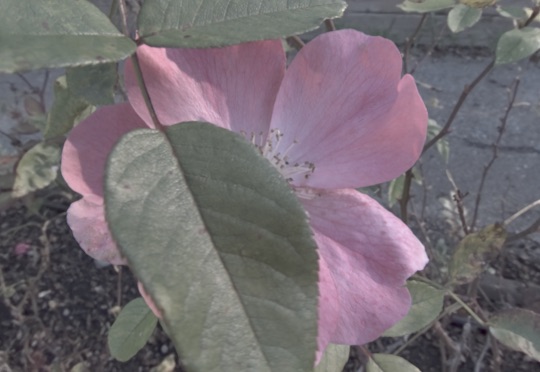} &
\includegraphics[width=0.23\linewidth,trim=240 132 190 130,clip=true]{nofcnn/7383_ours_y.jpg}
\\
\includegraphics[width=0.23\linewidth,trim=50 112 380 150,clip=true]{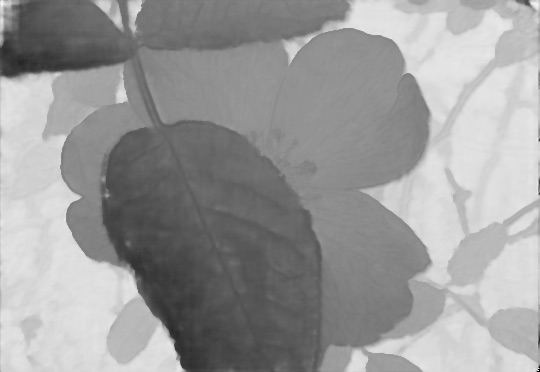} &
\includegraphics[width=0.23\linewidth,trim=240 132 190 130,clip=true]{nofcnn/7383_nof_d00.jpg} &
\includegraphics[width=0.23\linewidth,trim=50 112 380 150,clip=true]{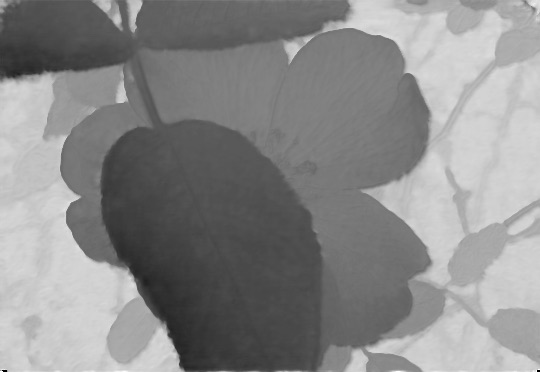} &
\includegraphics[width=0.23\linewidth,trim=240 132 190 130,clip=true]{nofcnn/7383_ours_d00.jpg}
\\
\includegraphics[width=0.23\linewidth,trim=50 112 380 150,clip=true]{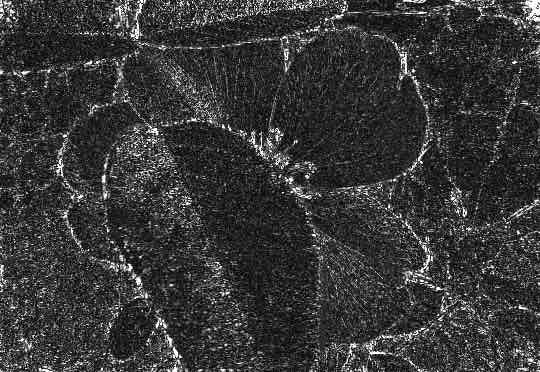} &
\includegraphics[width=0.23\linewidth,trim=240 132 190 130,clip=true]{nofcnn/7383_nof_error.jpg} &
\includegraphics[width=0.23\linewidth,trim=50 112 380 150,clip=true]{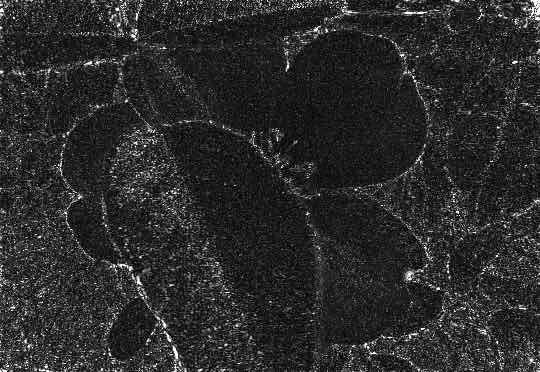} &
\includegraphics[width=0.23\linewidth,trim=240 132 190 130,clip=true]{nofcnn/7383_ours_error.jpg}
\end{tabular}
\caption{Influence of the features network. We show the reconstructed center view (top), the disparity map $d_{0,0}$ (middle) and the error image (bottom), which is clipped into the range $[0,0.04]$. We encourage the reader to look at the electronic version of this paper to better see the details.}
\label{fig:features} %Testing light field corresponds to {\it IMG\_7383\_eslf} from {\it Flowers}.
\end{figure}
%%%%%%%%%%%%%%%%%%%%%%%%%%%% 

\subsection{Effect of Using the Features CNN}
We now compare the proposed network against one that does not have a first stage for feature extraction and instead inputs to the disparity network are directly the light field views, as it is done for instance in~\cite{srinivasan2017learning}.
In this case, in the disparity CNN we included more convolutional layers to have the same number of trainable variables.
 In Table~\ref{tab:othermodels} we can see the gains in performance when using feature extraction, an average PSNR of $38.24$ opposed to $37.81$ when we do not use it in the {\it Flowers} test set. Moreover, on the {\it Diverse} test set it considerably decreases to $35.66$. A visual example is shown in Figure~\ref{fig:features}. Without the features network the method produces inaccuracies in the disparity map, which results in a loss of textures and higher reconstruction errors.

%%%%%%%%%%%%%%%%%%%%%%%%%%%%
\begin{table}[t!]\centering
\caption{Quantitative comparison with LBVS~\cite{kalantari2016learning} and 4DLF~\cite{srinivasan2017learning}. The dataset in parenthesis after each method indicates the used training set, where {\it F} stands for {\it Flowers} and {\it D} for {\it Diverse}.}
\resizebox{\columnwidth}{!}{
\begin{tabular}{|c|c|c|c|c|c|c|}  \cline{2-7}
\multicolumn{1}{c}{}& \multicolumn{3}{|c|}{\it Flowers} & \multicolumn{3}{c|}{\it Diverse}\\ \cline{2-7} \noalign{\smallskip}\hline 

Method      		&MAE	& PSNR  	& SSIM 		&MAE     	& PSNR  	& SSIM   \\ \hline \hline 
LBVS ({\it D}) 		& 1,374 	& 34,37	& 0,9625		& 1.053	& 36.13	& 0.9799\\\hline 
4DLF ({\it F}) 		& 2.998  	& 33.10  	& 0.9510 		& 3.859  	&  30.61 	& 0.9369\\\hline 
Proposed ({\it F}) 	&\bf0.878 	&\bf 38.29	& \bf 0.9778 	&\bf0.797 	&38.13 	& \bf 0.9849\\\hline 
Proposed ({\it D}) 	&  0.982  	& 37.34 	& 0.9733  		& 0.805	& \bf38.14&0.9846 \\\hline
\end{tabular}}
\label{tab:quantitative_comparison}
\end{table}
%%%%%%%%%%%%%%%%%%%%%%%%%%%%

%%%%%%%%%%%%%%%%%%%%%%%%%%%%
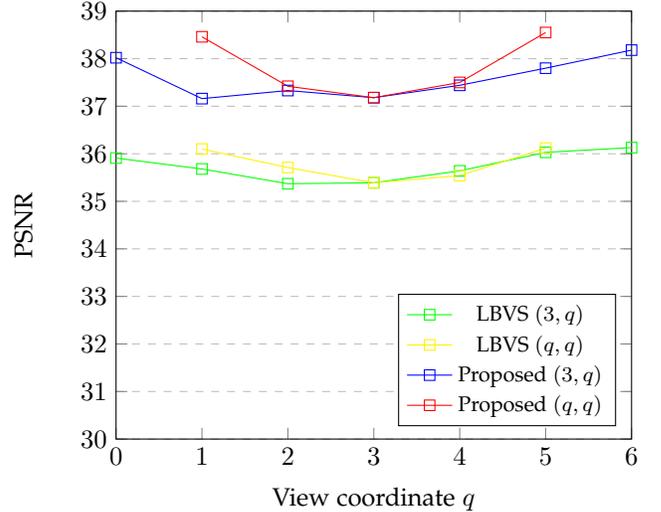
\begin{figure}[t!]\centering
\begin{tikzpicture}
\begin{axis}[
    xlabel={View coordinate $q$},
    ylabel={PSNR},
    xmin=0, xmax=6,
    ymin=30, ymax=39,
    xtick={0,1,2,3,4,5,6},
    ytick={30,31,32,33,34,35,36,37,38,39},
    legend pos=south east, %outer north east
    ymajorgrids=true,
    grid style=dashed,
]
\addplot[
    color=green,
    mark=square,
    ]
    coordinates {
	(0,35.91)(1,35.68)(2,35.37)(3,35.39)(4,35.64)(5,36.03)(6,36.13)
	};
    \addlegendentry{\footnotesize LBVS $(3, q)$}
\addplot[
    color=yellow,
    mark=square,
    ]
    coordinates {
    	(1,36.10)(2,35.71)(3,35.39)(4,35.54)(5,36.12)
	};
    \addlegendentry{\footnotesize LBVS $(q, q)$}
\addplot[
    color=blue,
    mark=square,
    ]
    coordinates {
    %(0,37.83)(1,37.68)(2,37.31)(3,37.28)(4,37.45)(5,37.68)(6,38.09) % Trained Flowers
    (0,38.02)(1,37.16)(2,37.33)(3,37.18)(4,37.44)(5,37.80)(6,38.18) % Trained Diverse
    };
    \addlegendentry{\footnotesize Proposed $(3, q)$}
\addplot[
    color=red,
    mark=square,
    ]
    coordinates {
	%(1,38.38)(2,37.39)(3,37.28)(4,37.33)(5,38.38) % Trained Flowers
	(1,38.46)(2,37.42)(3,37.18)(4,37.50)(5,38.55) % Trained Diverse
	};
    \addlegendentry{\footnotesize Proposed $(q, q)$}
\end{axis}
\end{tikzpicture}
\caption{Quality of the reconstruction for different views compared with LBVS~\cite{kalantari2016learning}. The evaluation is in terms of the PSNR, which is averaged over the {\it Diverse} test set for each particular view. For each method, we show the results for the row of views in the middle, with coordinates $(3,q)$, and the evaluation for the views of the form $(q,q)$, that lie in the diagonal that goes from the view $(0,0)$ to $(6,6)$. Both methods have been trained on the same {\it Diverse} training set.}
\label{fig:graphviewsK}
\end{figure}
%%%%%%%%%%%%%%%%%%%%%%%%%%%%

%%%%%%%%%%%%%%%%%%%%%%%%%%%%
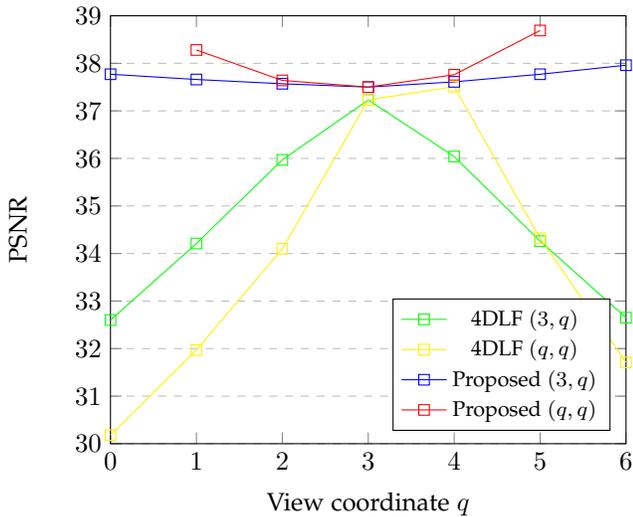
\begin{figure}[t!]\centering
\begin{tikzpicture}
\begin{axis}[
    xlabel={View coordinate $q$},
    ylabel={PSNR},
    xmin=0, xmax=6,
    ymin=30, ymax=39,
    xtick={0,1,2,3,4,5,6},
    ytick={30,31,32,33,34,35,36,37,38,39},
    legend pos=south east, %outer north east
    ymajorgrids=true,
    grid style=dashed,
]
\addplot[
    color=green,
    mark=square,
    ]
    coordinates {
	(0, 32.6)(1, 34.21)(2, 35.97)(3, 37.23)(4, 36.04)(5, 34.26)(6, 32.65)
	};
    \addlegendentry{\footnotesize 4DLF $(3, q)$}
\addplot[
    color=yellow,
    mark=square,
    ]
    coordinates {
	(0,30.18)(1,31.97)(2,34.10)(3, 37.23)(4,37.51)(5, 34.32)(6, 31.71)
	};
    \addlegendentry{\footnotesize 4DLF $(q, q)$}
\addplot[
    color=blue,
    mark=square,
    ]
    coordinates {
    (0,37.77)(1,37.66)(2,37.57)(3,37.50)(4,37.61)(5,37.77)(6,37.96)
    };
    \addlegendentry{\footnotesize Proposed $(3, q)$}
\addplot[
    color=red,
    mark=square,
    ]
    coordinates {
	(1, 38.28)(2, 37.64)(3, 37.50)(4, 37.76)(5, 38.69)
	};
    \addlegendentry{\footnotesize Proposed $(q, q)$}
\end{axis}
\end{tikzpicture}
\caption{Quality of the reconstruction for different views compared with 4DLF~\cite{srinivasan2017learning}. The evaluation is in terms of the PSNR, which is averaged over the {\it Flowers} test set for each particular view. For each method, we show the results for the row of views in the middle, with coordinates $(3,q)$, and the evaluation for the views of the form $(q,q)$, that lie in the diagonal that goes from the view $(0,0)$ to $(6,6)$. Both methods have been trained on the same {\it Flowers} training set.}
\label{fig:graphviewsS}
\end{figure}
%%%%%%%%%%%%%%%%%%%%%%%%%%%%

%%%%%%%%%%%%%%%%%%%%%%%%%%%% 
\begin{figure}[t!]\centering
\begin{tabular}{@{}c@{\hskip 0.2em}c@{\hskip 0.2em}c@{\hskip 0.2em}c@{}}
\multicolumn{2}{c}{LBVS} & \multicolumn{2}{c}{Proposed} 
\\
\includegraphics[width=0.24\linewidth]{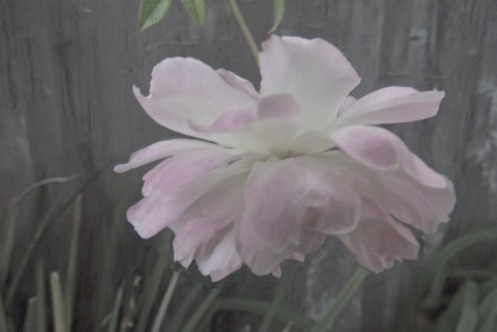} &
\includegraphics[width=0.24\linewidth]{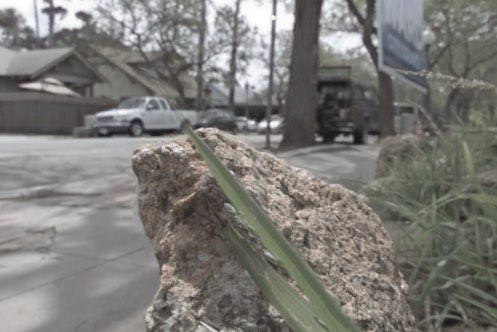} &
\includegraphics[width=0.24\linewidth]{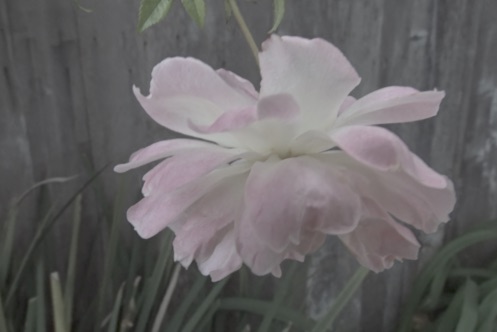} &
\includegraphics[width=0.24\linewidth]{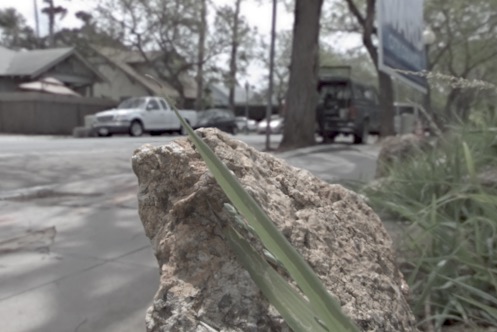} 
\\
\includegraphics[width=0.24\linewidth,trim=107 200 280 20,clip=true]{comparisonKS/IMG_5102_eslf/Kalantari_04_05_invertAdjustTone.jpg} &
\includegraphics[width=0.24\linewidth,trim=107 160 280 60,clip=true]{comparisonKS/Rock/Kalantari_04_05_invertAdjustTone.jpg} &
\includegraphics[width=0.24\linewidth,trim=107 200 280 20,clip=true]{comparisonKS/IMG_5102_eslf/y_crop.jpg} &
\includegraphics[width=0.24\linewidth,trim=107 160 280 60,clip=true]{comparisonKS/Rock/y_crop.jpg} 
\\
\includegraphics[width=0.24\linewidth]{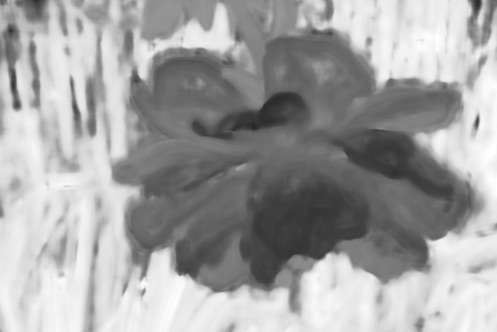} &
\includegraphics[width=0.24\linewidth]{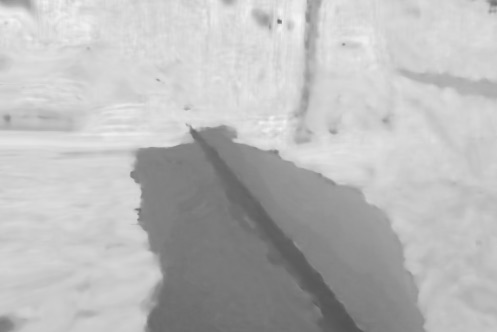} &
\includegraphics[width=0.24\linewidth]{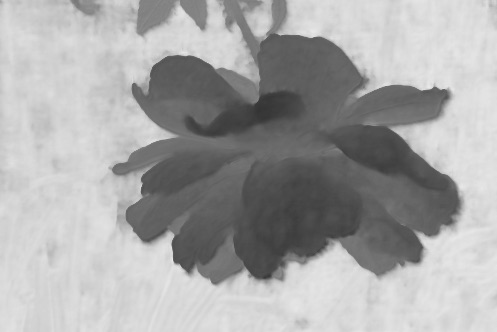} &
\includegraphics[width=0.24\linewidth]{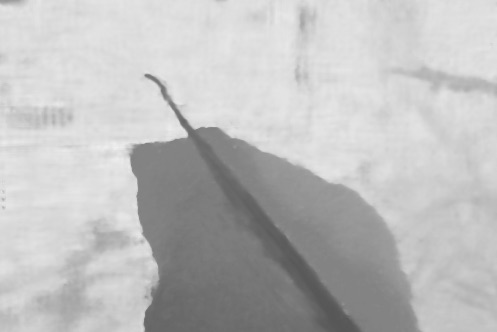} 
\\
\includegraphics[width=0.24\linewidth,trim=107 200 280 20,clip=true]{comparisonKS/IMG_5102_eslf/Kalantari_depth} &
\includegraphics[width=0.24\linewidth,trim=107 160 280 60,clip=true]{comparisonKS/Rock/Kalantari_depth_04_05_crop} &
\includegraphics[width=0.24\linewidth,trim=107 200 280 20,clip=true]{comparisonKS/IMG_5102_eslf/d00_crop} &
\includegraphics[width=0.24\linewidth,trim=107 160 280 60,clip=true]{comparisonKS/Rock/d00_crop} 
\\
\includegraphics[width=0.24\linewidth]{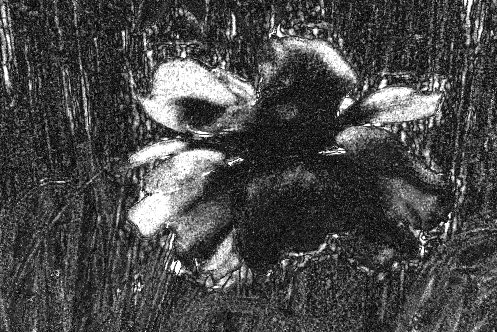} &
\includegraphics[width=0.24\linewidth]{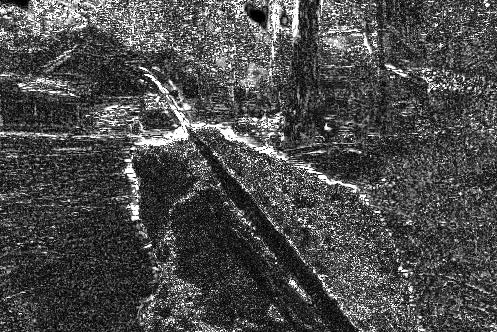} &
\includegraphics[width=0.24\linewidth]{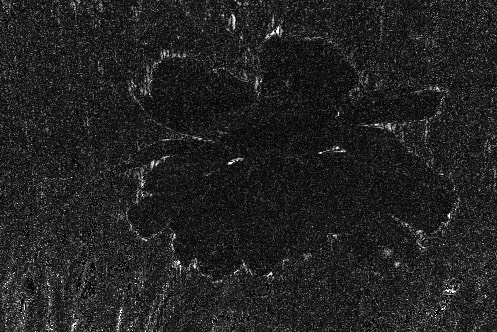} &
\includegraphics[width=0.24\linewidth]{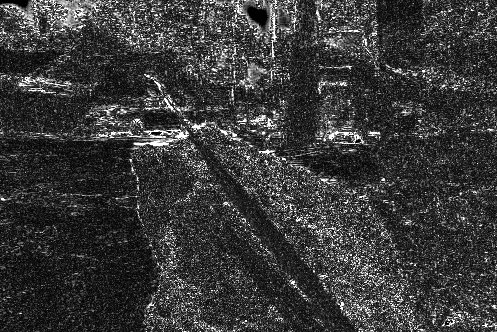} 
\\
\includegraphics[width=0.24\linewidth,trim=107 200 280 20,clip=true]{comparisonKS/IMG_5102_eslf/Kalantari_error} &
\includegraphics[width=0.24\linewidth,trim=107 160 280 60,clip=true]{comparisonKS/Rock/Kalantari_error_04_05} &
\includegraphics[width=0.24\linewidth,trim=107 200 280 20,clip=true]{comparisonKS/IMG_5102_eslf/error_crop} &
\includegraphics[width=0.24\linewidth,trim=107 160 280 60,clip=true]{comparisonKS/Rock/error_crop} 
\end{tabular}
\caption{Comparison against the state-of-the-art method LBVS~\cite{kalantari2016learning} on two examples, one from the {\it Flowers} test set (left) and the other from {\it Diverse} (right). For each method, we show the synthesized view (top), the estimated disparity map (middle) and the reconstruction error (bottom), which is clipped into $[0, 0.04]$.}
\label{fig:kalantari}
\end{figure}
%%%%%%%%%%%%%%%%%%%%%%%%%%%% 

%%%%%%%%%%%%%%%%%%%%%%%%%%%% 
\begin{figure}[t!]\centering
\begin{tabular}{@{}c@{\hskip 0.2em}c@{\hskip 0.2em}c@{\hskip 0.2em}c@{}}
\multicolumn{2}{c}{4DLF} & \multicolumn{2}{c}{Proposed} 
\\
\includegraphics[width=0.24\linewidth]{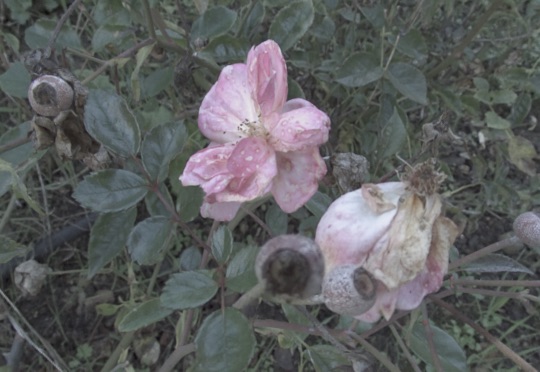} &
\includegraphics[width=0.24\linewidth]{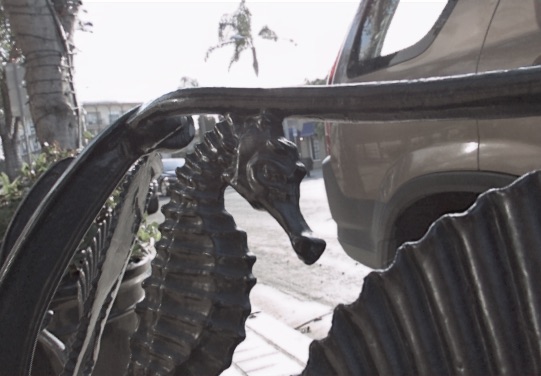} &
\includegraphics[width=0.24\linewidth]{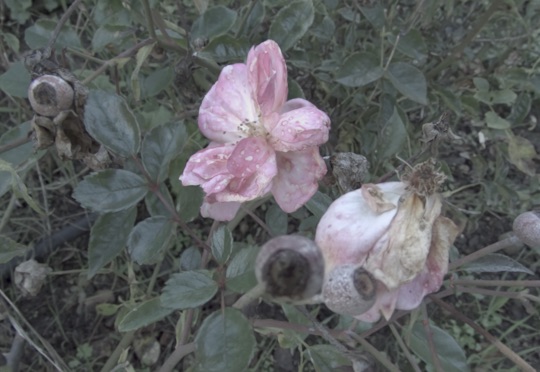} &
\includegraphics[width=0.24\linewidth]{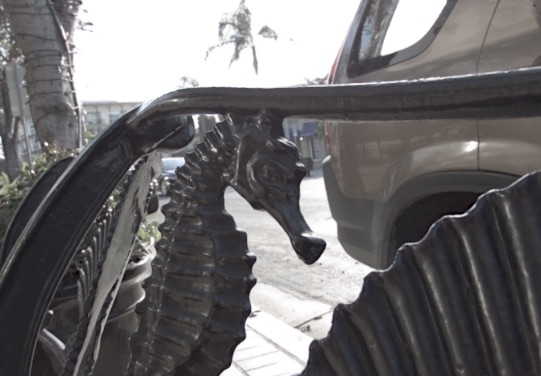} 
\\
\includegraphics[width=0.24\linewidth,trim=240 32 190 230,clip=true]{comparisonKS/IMG_7005_eslf/Srinivasan_y_p3_q4.jpg} &
\includegraphics[width=0.24\linewidth,trim=237 102 190 160,clip=true]{comparisonKS/Seahorse/Srinivasan_y_p3_q4.jpg} &
\includegraphics[width=0.24\linewidth,trim=240 32 190 230,clip=true]{comparisonKS/IMG_7005_eslf/ours_y.jpg} &
\includegraphics[width=0.24\linewidth,trim=237 102 190 160,clip=true]{comparisonKS/Seahorse/y.jpg} 
\\
\includegraphics[width=0.24\linewidth]{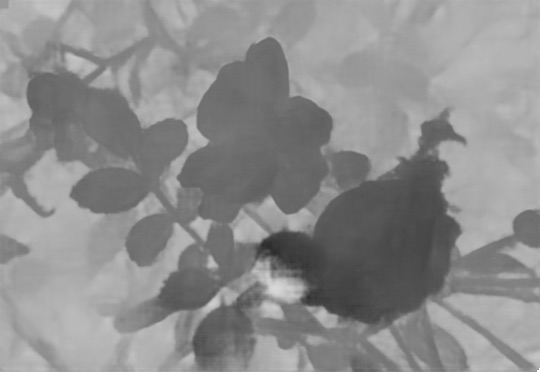} &
\includegraphics[width=0.24\linewidth]{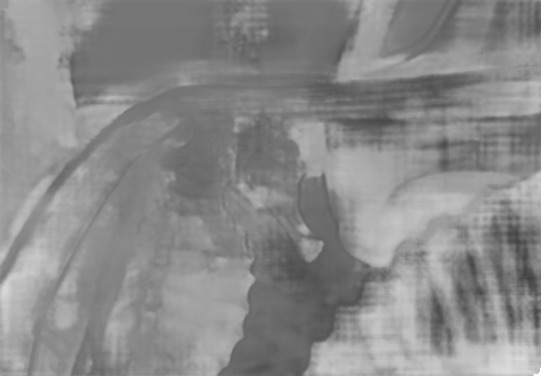} &
\includegraphics[width=0.24\linewidth]{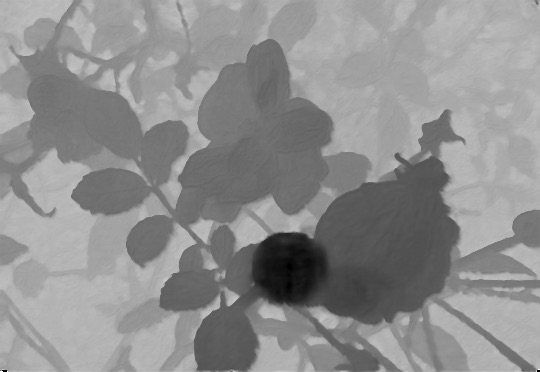} &
\includegraphics[width=0.24\linewidth]{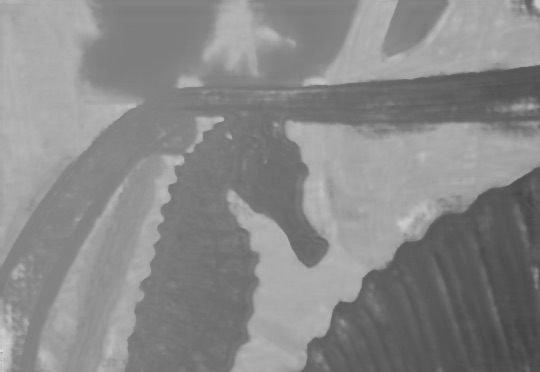} 
\\
\includegraphics[width=0.24\linewidth,trim=240 32 190 230,clip=true]{comparisonKS/IMG_7005_eslf/Srinivasan_depth_p3_q4.jpg} &
\includegraphics[width=0.24\linewidth,trim=237 102 190 160,clip=true]{comparisonKS/Seahorse/Srinivasan_depth_p3_q4.jpg} &
\includegraphics[width=0.24\linewidth,trim=240 32 190 230,clip=true]{comparisonKS/IMG_7005_eslf/ours_disparity_0_0.jpg} &
\includegraphics[width=0.24\linewidth,trim=237 102 190 160,clip=true]{comparisonKS/Seahorse/disparity_0_0.jpg} 
\\
\includegraphics[width=0.24\linewidth]{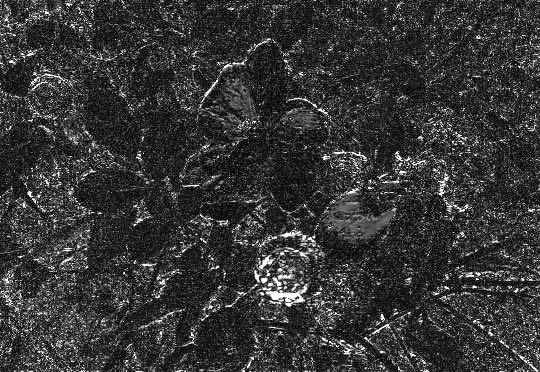} &
\includegraphics[width=0.24\linewidth]{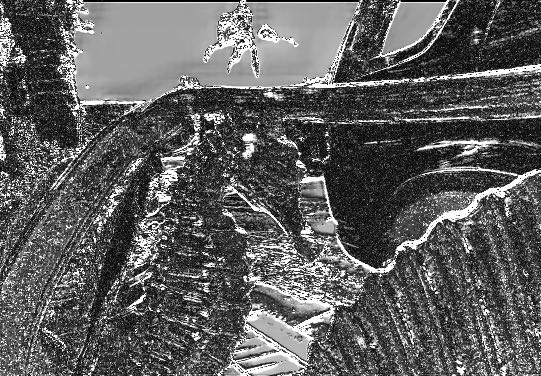}  &
\includegraphics[width=0.24\linewidth]{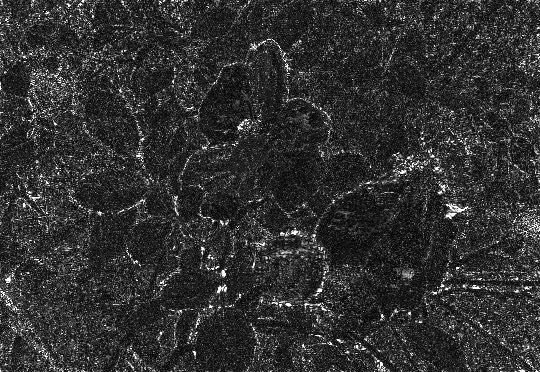} &
\includegraphics[width=0.24\linewidth]{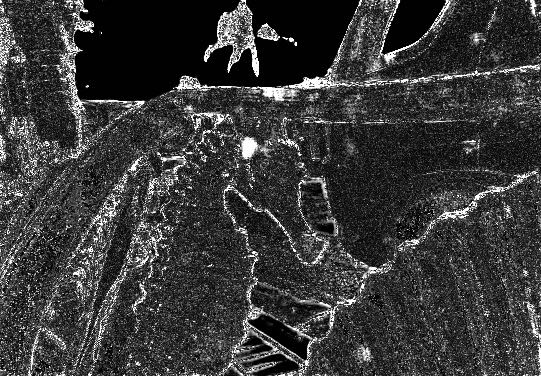} 
\\
\includegraphics[width=0.24\linewidth,trim=240 32 190 230,clip=true]{comparisonKS/IMG_7005_eslf/Srinivasan_error_p3_q4.jpg} &
\includegraphics[width=0.24\linewidth,trim=237 102 190 160,clip=true]{comparisonKS/Seahorse/Srinivasan_error_p3_q4.jpg}  &
\includegraphics[width=0.24\linewidth,trim=240 32 190 230,clip=true]{comparisonKS/IMG_7005_eslf/ours_error.jpg} &
\includegraphics[width=0.24\linewidth,trim=237 102 190 160,clip=true]{comparisonKS/Seahorse/error.jpg} 
\end{tabular}
\caption{Comparison against the state-of-the-art method 4DLF~\cite{srinivasan2017learning} on two examples, one from the {\it Flowers} test set (left) and the other from {\it Diverse} (right). For each method, we show the synthesized view (top), the estimated disparity map (middle) and the reconstruction error (bottom), which is clipped into $[0, 0.04]$.}
\label{fig:srinivasan}
\end{figure}
%%%%%%%%%%%%%%%%%%%%%%%%%%%% 

\subsection{Comparison against the State of the Art}
Table \ref{tab:quantitative_comparison} reports quantitative evaluation compared to the recent learning-based view synthesis from Kalantari et al.~\cite{kalantari2016learning} (LBVS), that reconstructs any view of the light field given the four corner ones; and the approach proposed by Srinivasan et al.~\cite{srinivasan2017learning} (4DLF), that reconstructs de 4D light field given the center view. LBVS has been trained on the {\it Diverse} training set, while 4DLF was optimized on the {\it Flowers} one. The evaluation metrics reported in the table have been averaged over the indicated test set and over the intersection of the views that have to be estimated for the three methods. According to the table, the proposed approach outperforms the other methods in all the metrics and in both datasets.
% LBVS the manual feature extraction is followed by a disparity and color networks. These two networks have a total of 1.64 million of parameters to train. 
% 4DLF has a disparity network and a residual one to refine the result at occlusions, which result in 2.07 million of trainable parameters.
% In our case, we have three networks for features, disparity and view selection which are defined with a total of 1.27 million of parameters.

In Figures~\ref{fig:graphviewsK} and \ref{fig:graphviewsS} we plot the PSNR as a function of the view postition $(p,q)$. In particular, we show the graph for the subset of views that are in the row of views in the middle of the light field, which have coordinates $(3,q)$, with $q\in\mathbb{Z}\cap[0,6]$; and the graph for the views of the form $(q,q)$, with $q\in\mathbb{Z}\cap[0,6]$, that lie in the diagonal that goes from the view $(0,0)$ to $(6,6)$. Figure~\ref{fig:graphviewsK} compares the proposed method with LBVS~\cite{kalantari2016learning}. Both models have been trained on the {\it Diverse} training set and, for each view position, the PSNR values have been averaged over the {\it Diverse} test set. We can observe that, for both methods, a higher PSNR is reported for those views that are closer to the input ones. However, LBVS values differ from our metrics in more than one point for all views. 

On the other hand, in Figure~\ref{fig:graphviewsS} we perform a similar comparison with 4DLF~\cite{srinivasan2017learning}. In this case, both models have been trained on the same {\it Flowers} training set and metrics are averaged over the {\it Flowers} test set for each angular position. According to the graph, the PSNR values of 4DLF drastically decrease as the novel view position distances from the input view, reaching values of almost 30. In our case, we also notice better results for those views that are closer to the input corner ones. However, differences in the PSNR values are smaller and always ranging between $37.50$ and $38.70$. 

Figure \ref{fig:kalantari} visually compares the result from LBVS~\cite{kalantari2016learning} with ours. This method uses the same estimated disparity map to warp each corner view. Therefore, disparity maps are less accurate at depth discontinuities than in our case. As it can be seen in both crops, their reconstruction has difficulties at occluded regions, resulting in blurred results and artifacts at these areas. Moreover, the method in LBVS synthesizes the new view using a CNN that outputs the novel color image. In some cases this may produce changes in colors, as it can be noticed in the flower. Besides, their method is unable to recover the tip of the leaf since disparity is not correctly estimated in this thin structure.

In Figure~\ref{fig:srinivasan} we compare our results with 4DLF~\cite{srinivasan2017learning}. Inferring a 4D light field from only one view may seem an advantage compared to our method. However, their method does not work properly with other images than flowers and it fails when dealing with complex scenes, where there is more than one object in the foreground. Although both methods have been trained on the same {\it Flowers} training set, their method is completely unable to model the geometry of the scene in some cases, resulting in high errors mostly located at object boundaries. On the contrary, our method better estimates disparity, which leads to smaller reconstruction errors.

%%%%%%%%%%%%%%%%%%%%%%%%%%%% 
\begin{figure}[t!]\centering
%\resizebox{\textwidth}{!}{
\begin{tabular}{@{}c@{\hskip 0.1em}c@{\hskip 0.1em}c@{\hskip 0.1em}c@{}}
Reference & Proposed & PBM & OADE
\\
\includegraphics[width=0.24\linewidth]{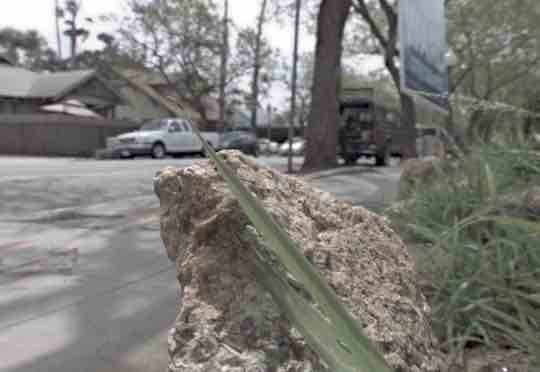} &
\includegraphics[width=0.24\linewidth]{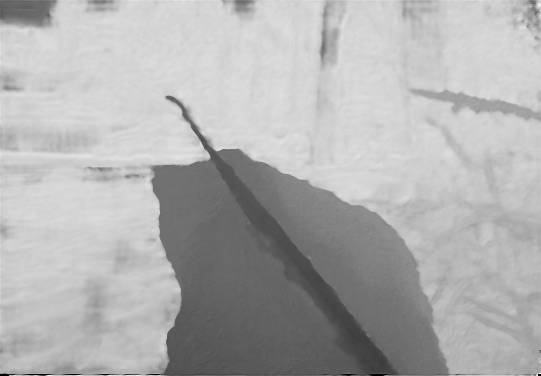} &
\includegraphics[width=0.24\linewidth]{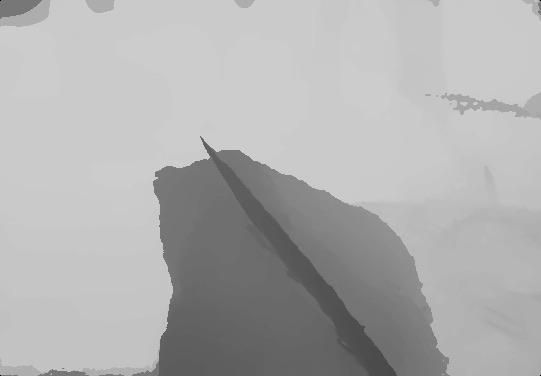} &
\includegraphics[width=0.24\linewidth]{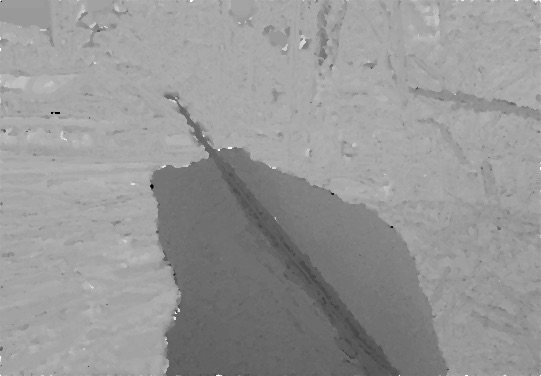} 
\\
\includegraphics[width=0.24\linewidth]{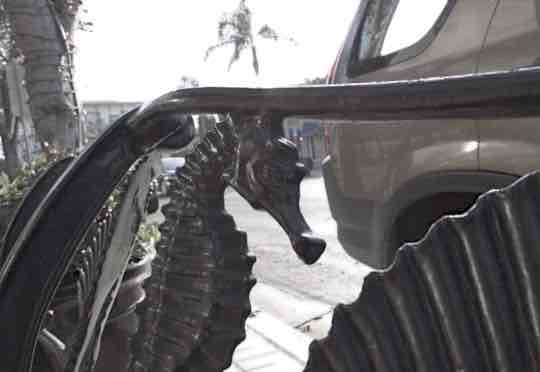} &
\includegraphics[width=0.24\linewidth]{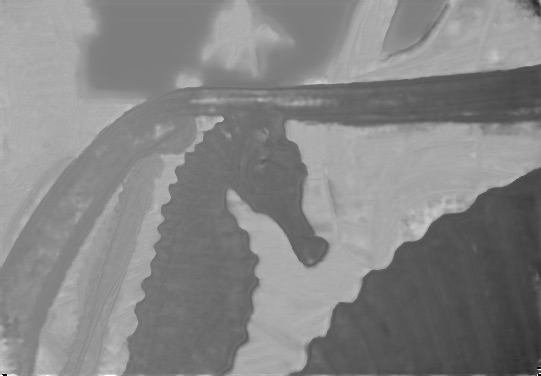} &
\includegraphics[width=0.24\linewidth]{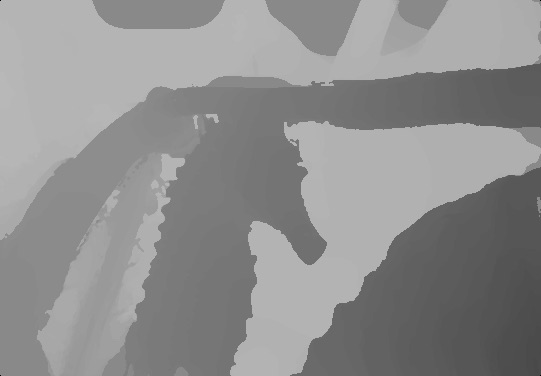} &
\includegraphics[width=0.24\linewidth]{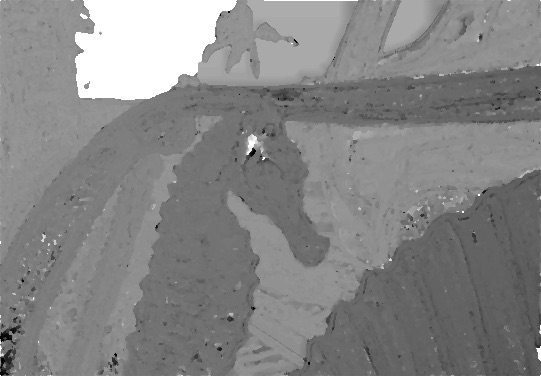} 
\end{tabular}%}
\caption{Estimated disparity compared to state-of-the-art methods PBM \cite{jeon2015accurate} and OADE \cite{wang2015occlusion}. In our case, disparity is estimated from the four corner views, while methods PBM and OADE make use of the complete 4D light field.}
\label{fig:depth_estimation}
\end{figure}
%%%%%%%%%%%%%%%%%%%%%%%%%%%% 

\subsection{Disparity Estimation}
Although the purpose of this work is view synthesis, the proposed method can also be used to estimate disparity in an unsupervised manner. In Figure~\ref{fig:depth_estimation} we qualitative compare our estimated disparity to state-of-the-art methods for depth estimation from light fields. Specifically, the phase-based method from Jeon et al.~\cite{jeon2015accurate} (PBM) and the occlusion-aware depth estimation from Wang et al.~\cite{wang2015occlusion} (OADE). To estimate disparity, these methods make use of the complete 4D light field, while our approach only uses the four corner views. In spite of that, our method shows to be competitive with respect to both PBM and OADE. %, being even more accurate at depth discontinuities and at fine objects. 

\subsection{Generalization} 
We assess the performance of the trained model tested on a different dataset by evaluating the model on the {\it Diverse}  test set. Furthermore, we trained our networks from scratch on the {\it Diverse} training set and tested on both {\it Flowers} and {\it Diverse} test sets. Evaluation metrics for these experiments are reported on Table~\ref{tab:quantitative_comparison}. Quantitatively, we can see how the method trained on {\it Flowers} yields a high PSNR also in the {\it Diverse} set. In addition, the training on the 100 images from the {\it Diverse} dataset yields similar performance than the one trained on {\it Flowers} on the testing images from the same dataset, while just a slightly worse performance is observed in the {\it Flowers} test set.

%%%%%%%%%%%%%%%%%%%%%%%%%%%%%%%%%%%%%%%%%%%%%%%%%%%%%%%
\begin{figure*}[t!]\centering
\begin{tabular}{@{}c@{\hskip 0.1em}c@{\hskip 0.1em}c@{\hskip 0.1em}c@{\hskip 0.1em}c@{\hskip 0.1em}c@{\hskip 0.1em}c@{\hskip 0.2em}c@{\hskip 0.1em}c@{\hskip 0.1em}c@{\hskip 0.1em}c@{\hskip 0.1em}c@{}} %input views
\multicolumn{4}{c}{Input}
\\
\multicolumn{2}{@{}c@{\hskip 0.1em}}{\begin{tikzpicture}
\node[above right, inner sep=0pt] (img) at (0,0) {\includegraphics[width=0.19\linewidth]{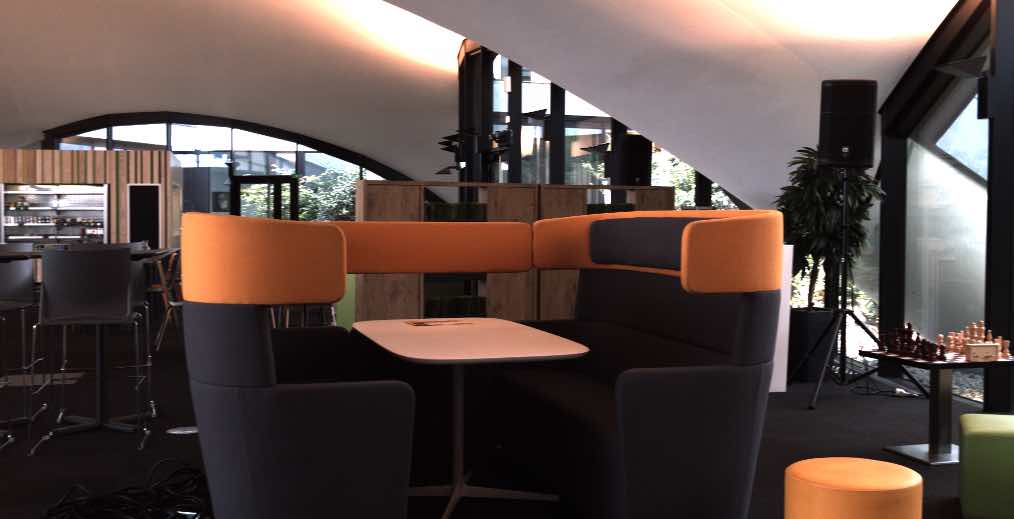}};
[overlay] \node at (0.5,0.35) {\color{white}$I_{0,0}$};
\end{tikzpicture}} &
\multicolumn{2}{@{}c@{}}{\begin{tikzpicture}
\node[above right, inner sep=0pt] (img) at (0,0) {\includegraphics[width=0.19\linewidth]{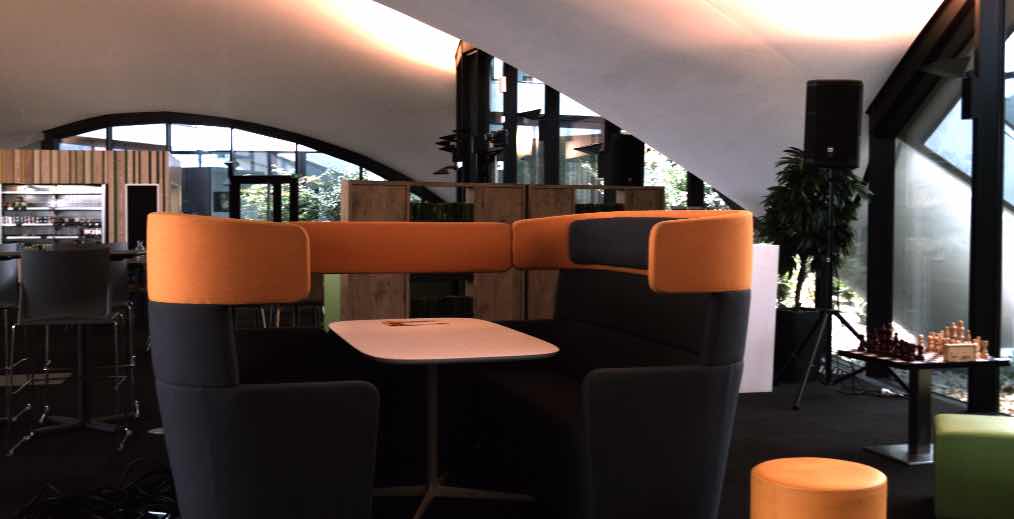}};
[overlay] \node at (0.5,0.35) {\color{white}$I_{0,2}$};
\end{tikzpicture}}
\\
\includegraphics[trim={150 190 694 159},clip=true,width=0.09\linewidth]{resultsRig/indoorRotation_41/i00.jpg}&
\includegraphics[trim={650 190 194 159},clip=true,width=0.09\linewidth]{resultsRig/indoorRotation_41/i00.jpg} &
\includegraphics[trim={150 190 694 159},clip=true,width=0.09\linewidth]{resultsRig/indoorRotation_41/i01.jpg} &
\includegraphics[trim={650 190 194 159},clip=true,width=0.09\linewidth]{resultsRig/indoorRotation_41/i01.jpg} &
\\
\multicolumn{2}{@{}c@{\hskip 0.1em}}{\begin{tikzpicture}
\node[above right, inner sep=0pt] (img) at (0,0) {\includegraphics[width=0.19\linewidth]{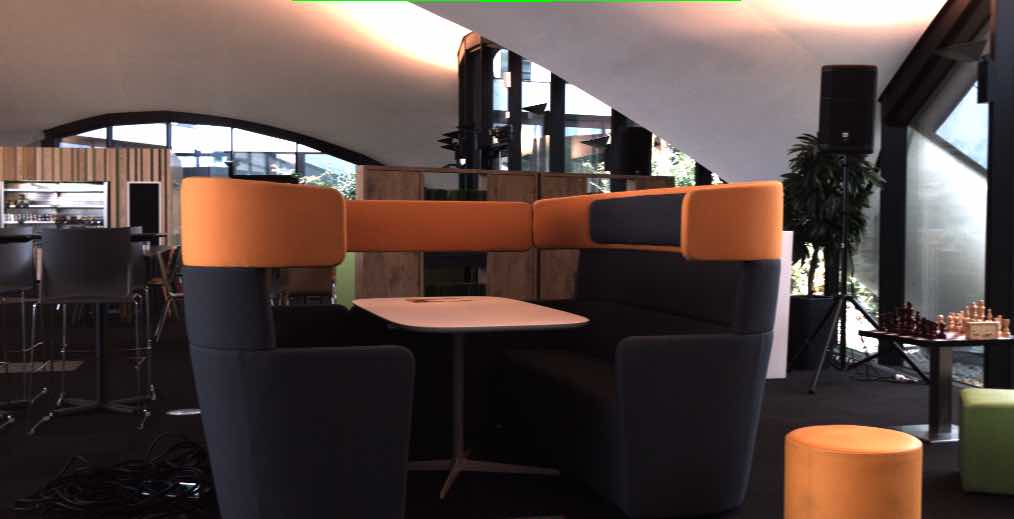}};
[overlay] \node at (0.5,0.35) {\color{white}$I_{2,0}$};
\end{tikzpicture}}&
\multicolumn{2}{@{}c@{}}{\begin{tikzpicture}
\node[above right, inner sep=0pt] (img) at (0,0) {\includegraphics[width=0.19\linewidth]{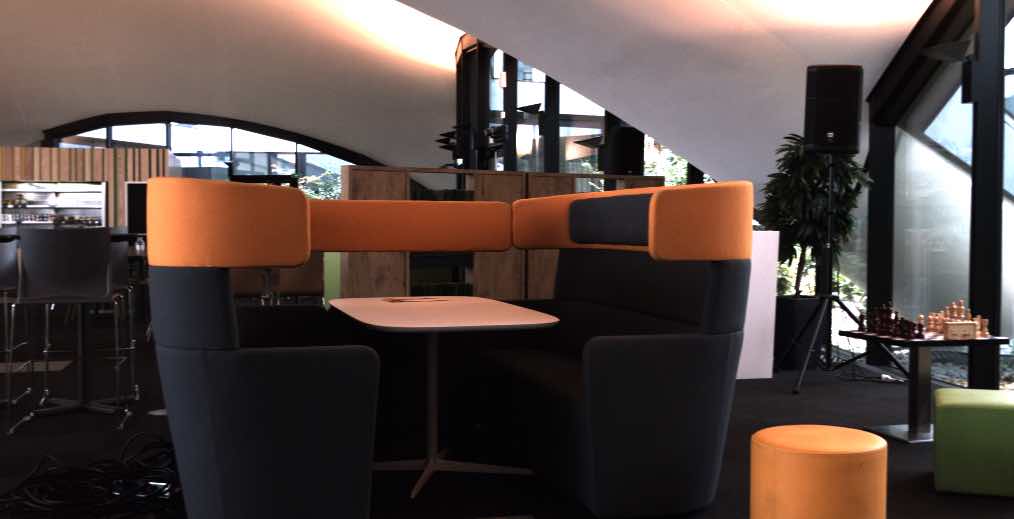}};
[overlay] \node at (0.5,0.35) {\color{white}$I_{2,2}$};
\end{tikzpicture}}
\\
\includegraphics[trim={150 190 694 159},clip=true,width=0.09\linewidth]{resultsRig/indoorRotation_41/i10.jpg} &
\includegraphics[trim={650 190 194 159},clip=true,width=0.09\linewidth]{resultsRig/indoorRotation_41/i10.jpg} &
\includegraphics[trim={150 190 694 159},clip=true,width=0.09\linewidth]{resultsRig/indoorRotation_41/i11.jpg} &
\includegraphics[trim={650 190 194 159},clip=true,width=0.09\linewidth]{resultsRig/indoorRotation_41/i11.jpg} &
\end{tabular}
\begin{tabular}{@{}c@{\hskip 0.2em}c@{\hskip 0.2em}c@{}}
Novel view $\hat{I}_{1,1}$
\\
\includegraphics[width=0.30\linewidth]{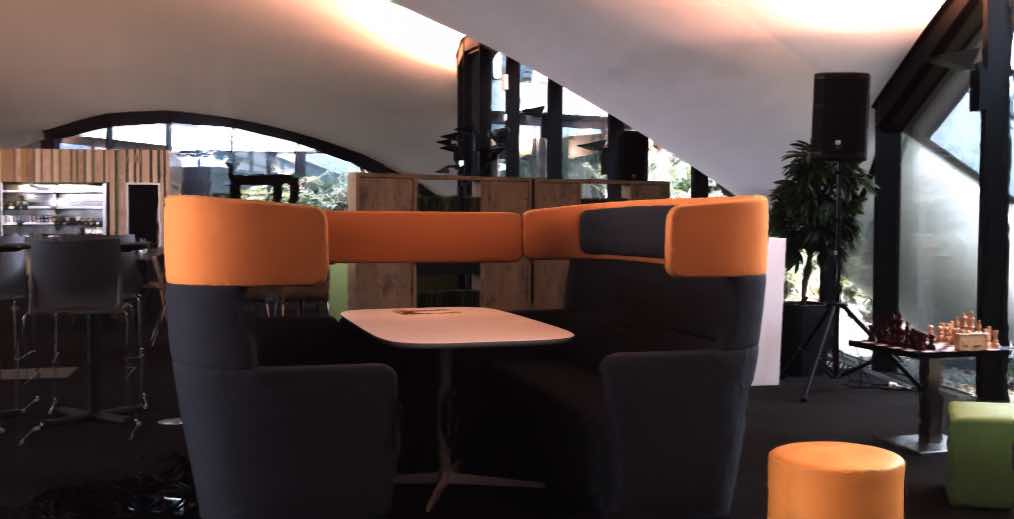} &
\includegraphics[trim={150 190 694 159},clip=true,width=0.13\linewidth]{resultsRig/indoorRotation_41/y.jpg} &
\includegraphics[trim={650 190 194 159},clip=true,width=0.13\linewidth]{resultsRig/indoorRotation_41/y.jpg} 
\\
Error
\\
\includegraphics[width=0.3\linewidth]{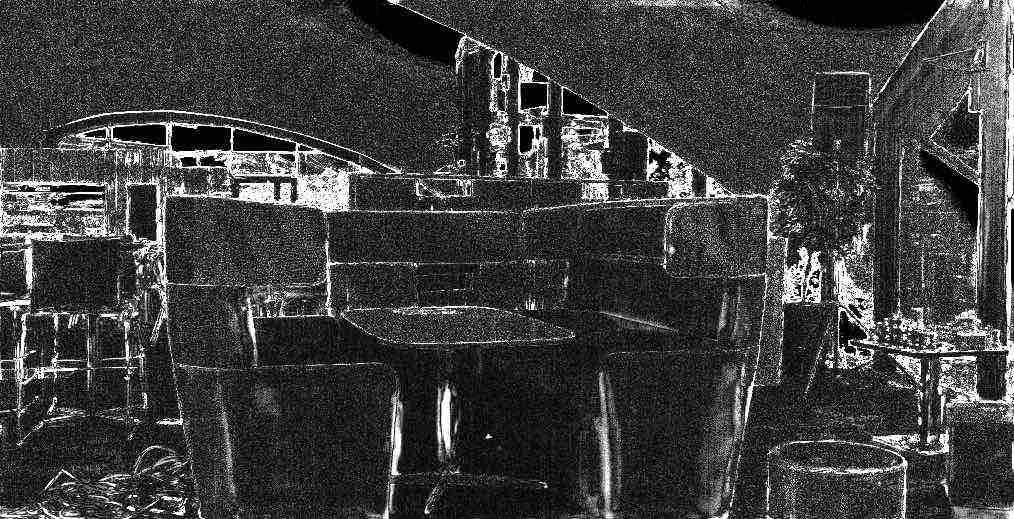} &
\includegraphics[trim={150 190 694 159},clip=true,width=0.13\linewidth]{resultsRig/indoorRotation_41/error.jpg} &
\includegraphics[trim={650 190 194 159},clip=true,width=0.13\linewidth]{resultsRig/indoorRotation_41/error.jpg} 
\end{tabular}

\medskip

%indoorTVscreevFar_111
\begin{tabular}{@{}c@{\hskip 0.1em}c@{\hskip 0.1em}c@{\hskip 0.1em}c@{\hskip 0.1em}c@{\hskip 0.1em}c@{\hskip 0.1em}c@{\hskip 0.2em}c@{\hskip 0.1em}c@{\hskip 0.1em}c@{\hskip 0.1em}c@{\hskip 0.1em}c@{}} %input views
\multicolumn{4}{c}{Input}
\\
\multicolumn{2}{@{}c@{\hskip 0.1em}}{\begin{tikzpicture}
\node[above right, inner sep=0pt] (img) at (0,0) {\includegraphics[width=0.19\linewidth]{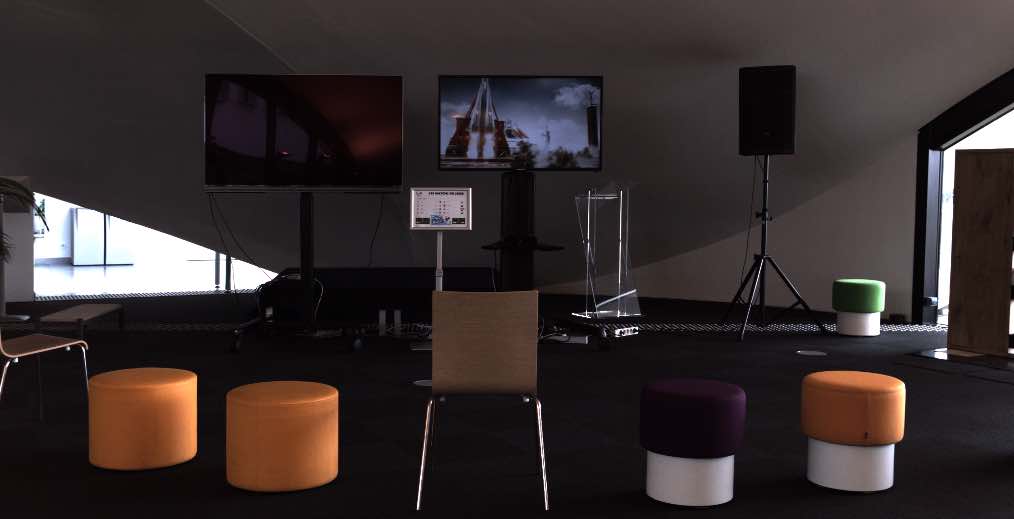}};
[overlay] \node at (0.5,0.35) {\color{white}$I_{0,0}$};
\end{tikzpicture}} &
\multicolumn{2}{@{}c@{}}{\begin{tikzpicture}
\node[above right, inner sep=0pt] (img) at (0,0) {\includegraphics[width=0.19\linewidth]{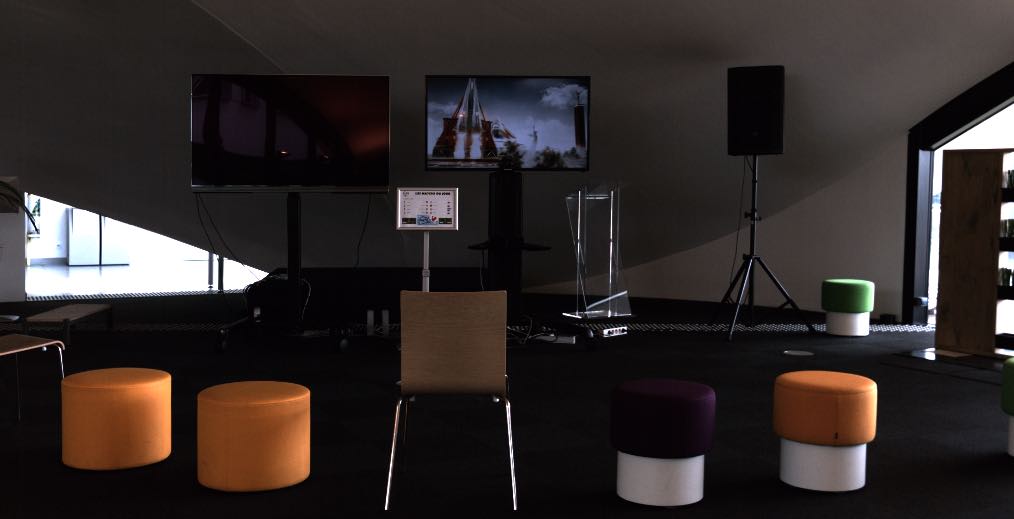}};
[overlay] \node at (0.5,0.35) {\color{white}$I_{0,2}$};
\end{tikzpicture}}
\\
\includegraphics[trim={150 190 694 159},clip=true,width=0.09\linewidth]{resultsRig/indoorTVscreevFar_111/i00.jpg}&
\includegraphics[trim={650 190 194 159},clip=true,width=0.09\linewidth]{resultsRig/indoorTVscreevFar_111/i00.jpg} &
\includegraphics[trim={150 190 694 159},clip=true,width=0.09\linewidth]{resultsRig/indoorTVscreevFar_111/i01.jpg} &
\includegraphics[trim={650 190 194 159},clip=true,width=0.09\linewidth]{resultsRig/indoorTVscreevFar_111/i01.jpg} &
\\
\multicolumn{2}{@{}c@{\hskip 0.1em}}{\begin{tikzpicture}
\node[above right, inner sep=0pt] (img) at (0,0) {\includegraphics[width=0.19\linewidth]{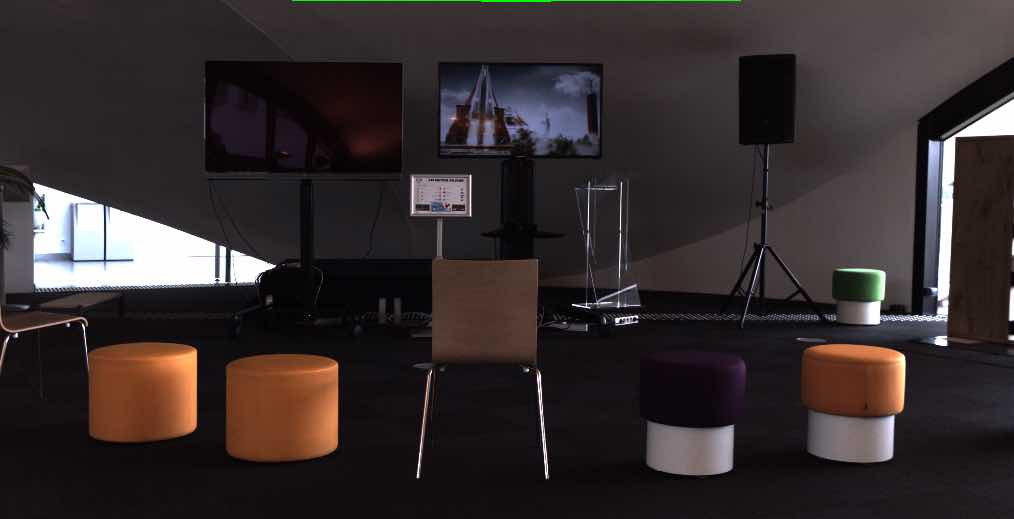}};
[overlay] \node at (0.5,0.35) {\color{white}$I_{2,0}$};
\end{tikzpicture}}&
\multicolumn{2}{@{}c@{}}{\begin{tikzpicture}
\node[above right, inner sep=0pt] (img) at (0,0) {\includegraphics[width=0.19\linewidth]{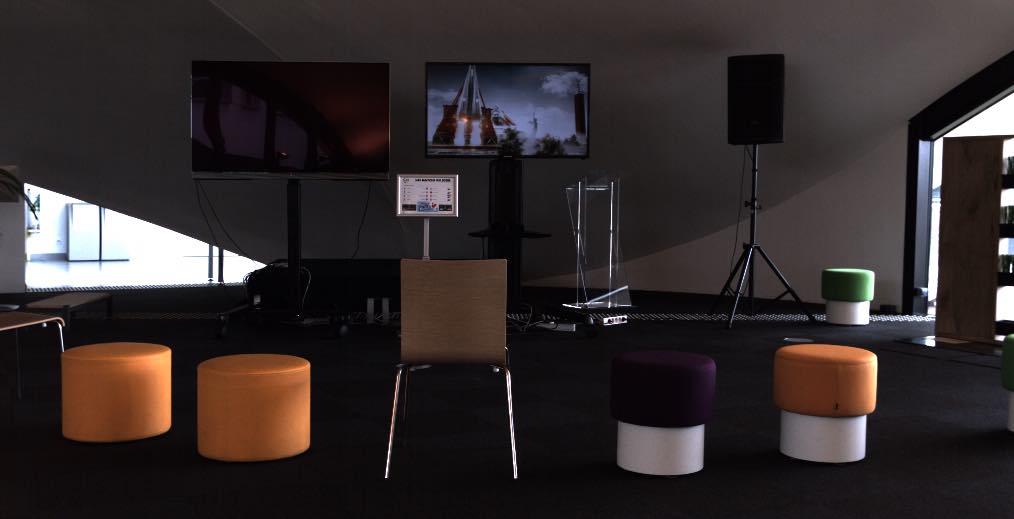}};
[overlay] \node at (0.5,0.35) {\color{white}$I_{2,2}$};
\end{tikzpicture}}
\\
\includegraphics[trim={150 190 694 159},clip=true,width=0.09\linewidth]{resultsRig/indoorTVscreevFar_111/i10.jpg} &
\includegraphics[trim={650 190 194 159},clip=true,width=0.09\linewidth]{resultsRig/indoorTVscreevFar_111/i10.jpg} &
\includegraphics[trim={150 190 694 159},clip=true,width=0.09\linewidth]{resultsRig/indoorTVscreevFar_111/i11.jpg} &
\includegraphics[trim={650 190 194 159},clip=true,width=0.09\linewidth]{resultsRig/indoorTVscreevFar_111/i11.jpg} &
\end{tabular}
\begin{tabular}{@{}c@{\hskip 0.2em}c@{\hskip 0.2em}c@{}}
Novel view $\hat{I}_{1,1}$
\\
\includegraphics[width=0.30\linewidth]{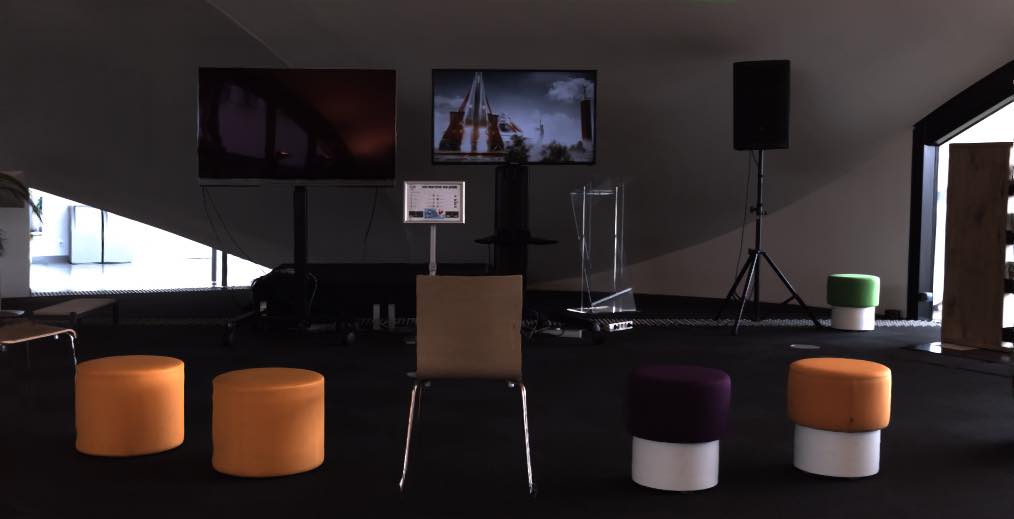} &
\includegraphics[trim={150 190 694 159},clip=true,width=0.13\linewidth]{resultsRig/indoorTVscreevFar_111/y.jpg} &
\includegraphics[trim={650 190 194 159},clip=true,width=0.13\linewidth]{resultsRig/indoorTVscreevFar_111/y.jpg} 
\\
Error
\\
\includegraphics[width=0.3\linewidth]{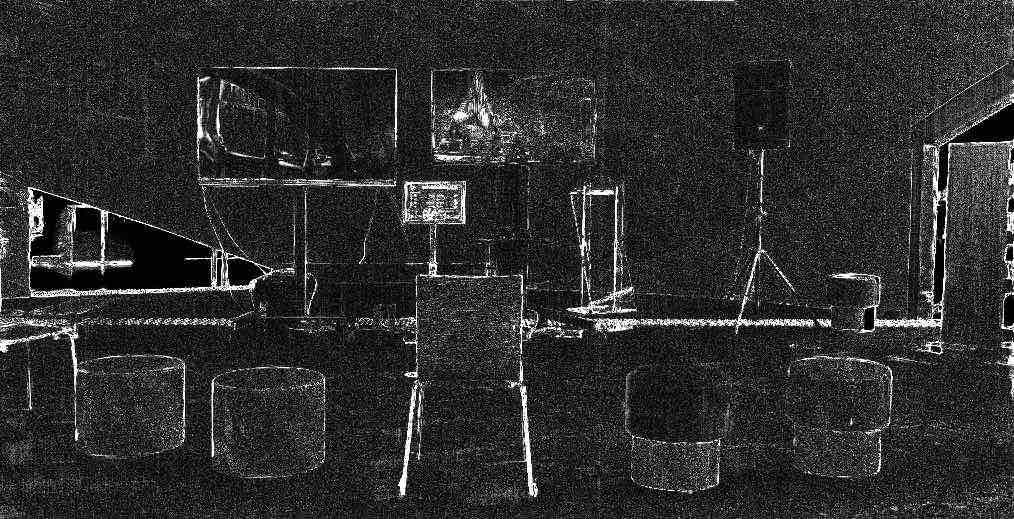} &
\includegraphics[trim={150 190 694 159},clip=true,width=0.13\linewidth]{resultsRig/indoorTVscreevFar_111/error.jpg} &
\includegraphics[trim={650 190 194 159},clip=true,width=0.13\linewidth]{resultsRig/indoorTVscreevFar_111/error.jpg} 
\end{tabular}

\medskip

%openSpaceEtage_231
\begin{tabular}{@{}c@{\hskip 0.1em}c@{\hskip 0.1em}c@{\hskip 0.1em}c@{\hskip 0.1em}c@{\hskip 0.1em}c@{\hskip 0.1em}c@{\hskip 0.2em}c@{\hskip 0.1em}c@{\hskip 0.1em}c@{\hskip 0.1em}c@{\hskip 0.1em}c@{}} %input views
\multicolumn{4}{c}{Input}
\\
\multicolumn{2}{@{}c@{\hskip 0.1em}}{\begin{tikzpicture}
\node[above right, inner sep=0pt] (img) at (0,0) {\includegraphics[width=0.19\linewidth]{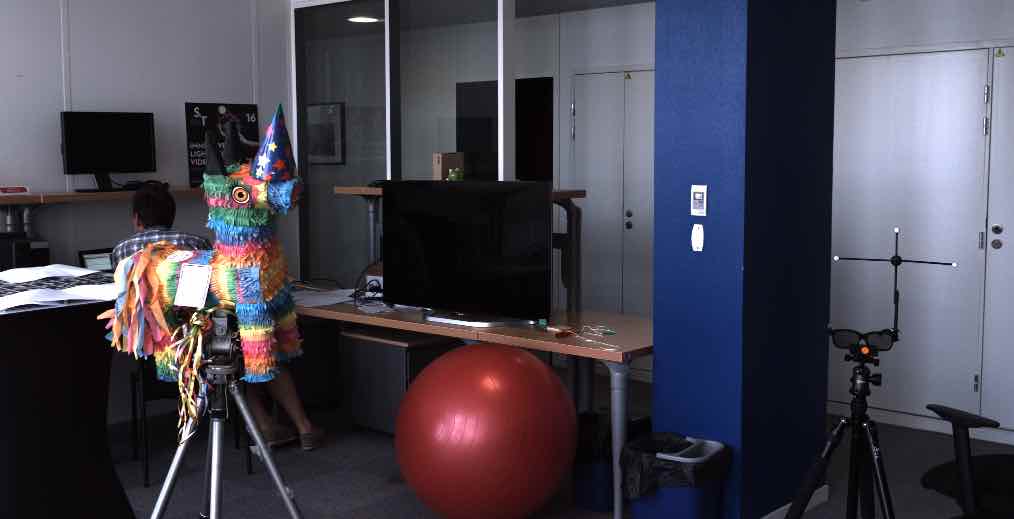}};
[overlay] \node at (0.5,0.35) {\color{white}$I_{0,0}$};
\end{tikzpicture}} &
\multicolumn{2}{@{}c@{}}{\begin{tikzpicture}
\node[above right, inner sep=0pt] (img) at (0,0) {\includegraphics[width=0.19\linewidth]{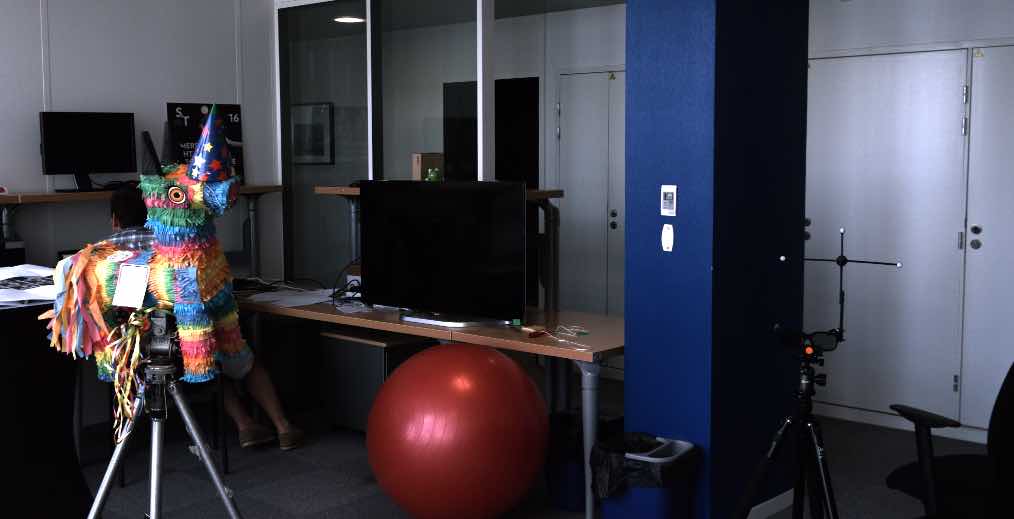}};
[overlay] \node at (0.5,0.35) {\color{white}$I_{0,2}$};
\end{tikzpicture}}
\\
\includegraphics[trim={150 190 694 159},clip=true,width=0.09\linewidth]{resultsRig/openSpaceEtage_231/i00.jpg}&
\includegraphics[trim={650 190 194 159},clip=true,width=0.09\linewidth]{resultsRig/openSpaceEtage_231/i00.jpg} &
\includegraphics[trim={150 190 694 159},clip=true,width=0.09\linewidth]{resultsRig/openSpaceEtage_231/i01.jpg} &
\includegraphics[trim={650 190 194 159},clip=true,width=0.09\linewidth]{resultsRig/openSpaceEtage_231/i01.jpg} &
\\
\multicolumn{2}{@{}c@{\hskip 0.1em}}{\begin{tikzpicture}
\node[above right, inner sep=0pt] (img) at (0,0) {\includegraphics[width=0.19\linewidth]{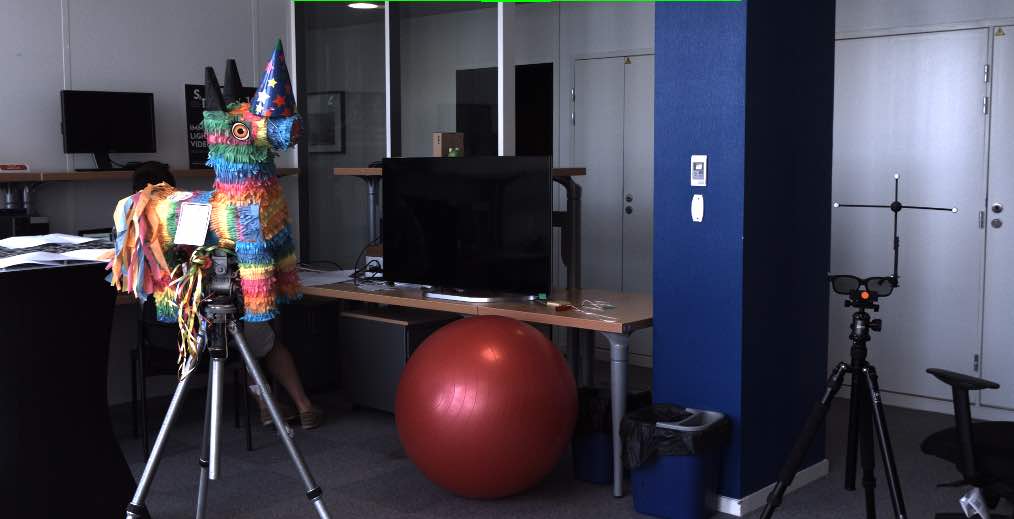}};
[overlay] \node at (0.5,0.35) {\color{white}$I_{2,0}$};
\end{tikzpicture}}&
\multicolumn{2}{@{}c@{}}{\begin{tikzpicture}
\node[above right, inner sep=0pt] (img) at (0,0) {\includegraphics[width=0.19\linewidth]{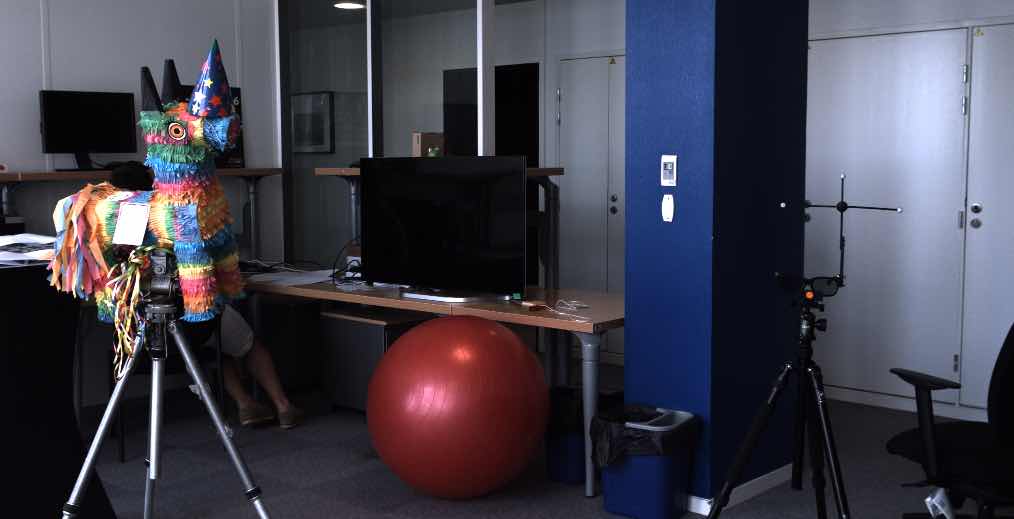}};
[overlay] \node at (0.5,0.35) {\color{white}$I_{2,2}$};
\end{tikzpicture}}
\\
\includegraphics[trim={150 190 694 159},clip=true,width=0.09\linewidth]{resultsRig/openSpaceEtage_231/i10.jpg} &
\includegraphics[trim={650 190 194 159},clip=true,width=0.09\linewidth]{resultsRig/openSpaceEtage_231/i10.jpg} &
\includegraphics[trim={150 190 694 159},clip=true,width=0.09\linewidth]{resultsRig/openSpaceEtage_231/i11.jpg} &
\includegraphics[trim={650 190 194 159},clip=true,width=0.09\linewidth]{resultsRig/openSpaceEtage_231/i11.jpg} &
\end{tabular}
\begin{tabular}{@{}c@{\hskip 0.2em}c@{\hskip 0.2em}c@{}}
Novel view $\hat{I}_{1,1}$
\\
\includegraphics[width=0.30\linewidth]{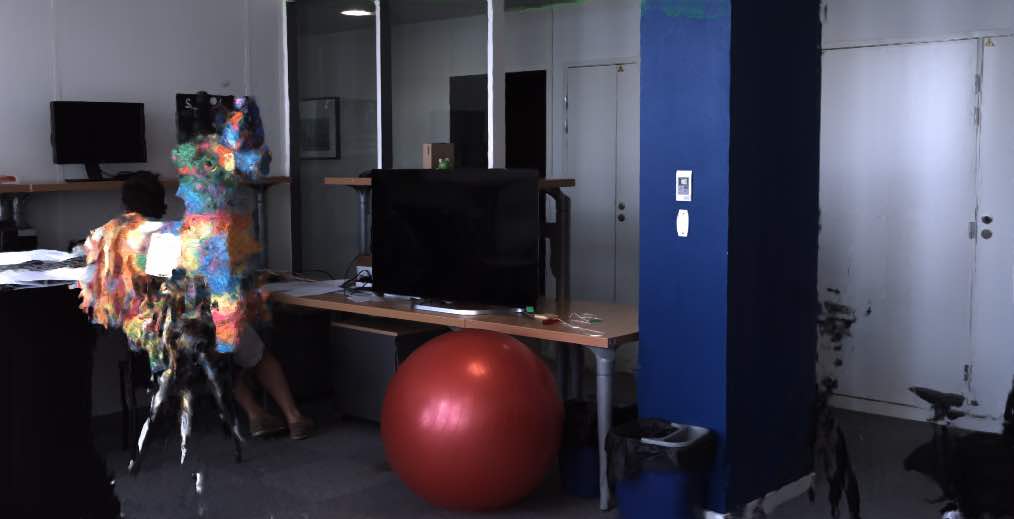} &
\includegraphics[trim={150 190 694 159},clip=true,width=0.13\linewidth]{resultsRig/openSpaceEtage_231/y.jpg} &
\includegraphics[trim={650 190 194 159},clip=true,width=0.13\linewidth]{resultsRig/openSpaceEtage_231/y.jpg} 
\\
Error
\\
\includegraphics[width=0.3\linewidth]{resultsRig/indoorTVscreevFar_111/error.jpg} &
\includegraphics[trim={150 190 694 159},clip=true,width=0.13\linewidth]{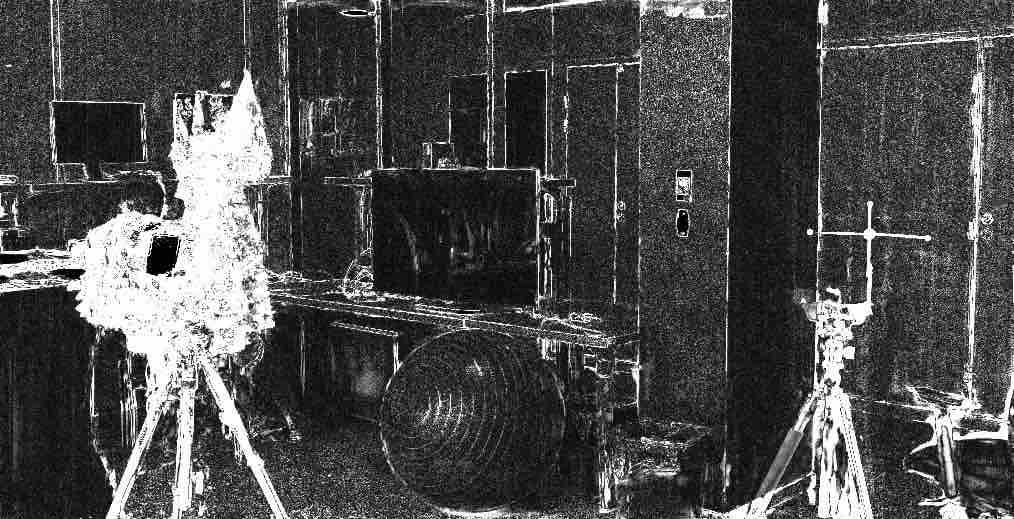} &
\includegraphics[trim={650 190 194 159},clip=true,width=0.13\linewidth]{resultsRig/openSpaceEtage_231/error.jpg} 
\end{tabular}
\caption{Visual results on light fields acquired with the camera rig presented in~\cite{sabater2017dataset}. Error images are clipped into the range $[0,0.04]$ for images in $[0,1]$.}
\label{fig:rigResults}
\end{figure*}
%%%%%%%%%%%%%%%%%%%%%%%%%%%%%%%%%%%%%%%%%%%%%%%%%%%%%%%

\subsection{Wide-Baseline Light Fields}
Wide-baseline light fields make more difficult the problem of view synthesis than with Lytro images. Wider baselines involve having larger disparities and therefore much larger occluded areas. The proposed method specially treats the occlusion problem by assuming differences in the disparity maps and computing four different ones for each corner view. Moreover, we have seen that the selection network is able to detect occluded pixels and discard inaccurate reconstructions on these parts. 

In this section, we apply the proposed approach to wide-baseline light fields. In particular, to light field images captured with the camera rig presented in~\cite{sabater2017dataset}. The baseline between two consecutive cameras of this rig is $7$cm. In this complex case we have to deal with larger disparities than with Lytro light fields. Therefore, we cannot directly use the same networks as the ones used in the previous case, since the receptive field of the disparity network will not be enough to match distant pixels. In the following we adapt the proposed networks to deal with this challenging case.

\subsubsection{Adaptation to Wide-Baseline Light Fields}
To increase the receptive field and provide the disparity CNN of more global information without introducing many parameters, we apply the same features CNN $f_e$ as before at three different dilation rates for the first five convolutional layers. These dilations are 2, 4 and 8, respectively. The output volumes given from each dilation rate are concatenated along the depth dimension and fused by means of $1\times1$ convolutions to obtain a 32-channel feature volume. 

The features CNN outputs a volume for each view $F_{0,0}, F_{0,N}, F_{N,0}$ and $F_{N,N}$. In the previous case, these features were concatenated and were the input to the disparity CNN. Here, as disparities and occluded areas are too large and trying to find correspondences between the four images at the same time might be too difficult for the network, we propose to compute disparity maps from horizontal and vertical pairs of views separately and then to fuse these disparity estimations by means of a simple convolutional network. With this strategy the disparity network can better establish matches between input images since the overlapping between horizontal or vertical pairs is larger than if we consider the four images at the same time.

Then, horizontal disparities are computed from the concatenation of horizontal pairs of views,
\begin{align}
(d_{0,0}^h, d_{0,N}^h) &= f_{d_h}(P, Q, F_{0,0}, F_{0,N}), \\
(d_{N,0}^h, d_{N,N}^h) &= f_{d_h}(P, Q, F_{N,0}, F_{N,N});
\end{align}
while vertical ones are computed from vertical pairs,
\begin{align}
(d_{0,0}^v, d_{N,0}^v) &= f_{d_v}(P, Q, F_{0,0}, F_{N,0}), \\
(d_{0,N}^v, d_{N,N}^v) &= f_{d_v}(P, Q, F_{0,N}, F_{N,N}).
\end{align}
Functions $f_{d_h}$ and $f_{d_v}$ are convolutional networks that have the same architecture as the disparity CNN ($f_{d}$) but replacing input and output sizes with 64 and 2 channels, respectively. 

Disparities estimated from horizontal and vertical displacements are fused into a single disparity map for each view, 
\begin{equation}
d_{s,t} = f_{d_f}(d_{s,t}^h, d_{s,t}^v),\quad s,t\in\{0,N\}
\end{equation}
with $f_{d_f}$ being a convolutional neural network of two layers with kernels $3\times3$ and $1\times1$, respectively. 

Once we have computed the four disparity maps, the algorithm follows as before. The four views are warped using the corresponding disparity according to Equation~\eqref{eq:warpings}. Then, images $P$ and $Q$, the warped views $I_{s,t}^w$, with $s,t\in\{0,N\}$ and disparity maps $d_{s,t}$, with  $s,t\in\{0,N\}$ are concatenated along the channel dimension and are the input to the selection network $f_s$. The selection network is exactly the same as in the Lytro case. Finally, the predicted center view is computed as a weighted average of the four warped views using as weights the selection masks, according to Equation~\eqref{eq:reconstruction}.

The receptive field of the network proposed for the wide-baseline case is $170$ pixels, compared to $97$ pixels for the plenoptic version. The full model has a total of $2.02$ million of parameters to learn. As training loss function, we use the one from Equation~\eqref{eq:loss} with two additional terms that enforce the warped views with the horizontal and vertical disparities to be similar to the ground truth image.

\subsubsection{Training details}
We train these networks from scratch on light field images captured with the camera rig presented in~\cite{sabater2017dataset}. Video sequences from indoor and outdoor scenarios have been recorded and one of every ten frames has been selected as training light field. These light fields have been rectified and viewpoints are arranged on a regular grid. From the available $4\times4$ views, we randomly select an array of $3\times3$ and, from the four corner images, we train the networks to estimate the center one. Also, these light fields have a spatial resolution of $2048\times1088$ and are spatially downsampled by a factor of $2$.

The training set contains $212$ light fields and, considering that we take subsets of $3\times3$ views, this results in a total of $848$ examples. We randomly extract patches of $250\times250$ from the training images to train the model. The network is optimized using the ADAM solver~\cite{kingma2014adam} with $\beta_1=0.9, \beta_2=0.999, \epsilon=10^{-8}$, a learning rate of $0.001$ and a batch size of $1$. Weights are randomly initialized using the Xavier method~\cite{glorot2010understanding} and the softmax $\beta$ is initialized to $1$. The maximum disparity has been set to $d_{\text{max}}=60$. The method converges after $300$ thousand iterations, which approximately takes 2 days and 20 hours on a GeForce GTX 1080 Ti GPU. At test time, it takes $20$ seconds to synthesize a new instance.

\subsubsection{Application to Wide-Baseline Light Fields}
Figure~\ref{fig:rigResults} illustrates an example of view synthesis for different light fields captured with the camera rig presented in~\cite{sabater2017dataset}, using the proposed method. The method shows very promising results as it can be seen in the figure. The crops from the input views give an intuition of how large are occlusions in each case. By looking at the error images, in general most of these occluded parts do not present large errors. In the first and second examples, largest errors are present in some parts of the background, mainly on bright areas. 

In the last example, we have two objects with large disparities and the method is unable to correctly estimate them. This results in a blurred reconstruction and a thin structure that does not appear in the predicted center view. This suggests that the receptive field of the network is not enough to deal with these large disparities. However, the method that was first designed to cope with plenoptic cameras generally yields promising results for this challenging case, being able to detect from each view that parts that are visible in the center one.

%%%%%%%%%%%%%%%%%%%%%%%%%%%%%%%%%%%%%%%%%%
%%															  %%
%%						CONCLUSIONS				        	           %%
%%														 	  %%
%%%%%%%%%%%%%%%%%%%%%%%%%%%%%%%%%%%%%%%%%%
\section{Conclusions}\label{sec:conclusions}
In this work, we proposed a novel learning-based approach for new view synthesis for light field images. In particular, given the four corner views of a light field, we have tackled the problem of estimating any view in between. The method uses three sequential networks for feature extraction, for disparity estimation and another for view selection. 
%First, features are extracted from each input view using a convolutional network that shares weights across all views.
%These features are the input to a disparity network that estimates disparity between the center view and each corner view. 
Compared to the state of the art, we propose to compute four different disparity maps in order to deal with the occlusions problem. %The selection network outputs a selection mask for each view. These masks determine the contribution of each view on the synthesized result. This network is able to detect occluded areas and discard them for computing the novel view. 
Experiments have demonstrated the importance of using this strategy, jointly with the selection network, to obtain accurate results at occlusions. The method has proved to outperform the state of the art for Lytro light fields and its application to light fields from the camera rig from~\cite{sabater2017dataset} has given very promising results. 

As future work, we plan to focus on wide-baseline light fields and work on architectures for this special case, in which
networks should incorporate more context information in order to deal with large disparities and occlusions.

% use section* for acknowledgment
\ifCLASSOPTIONcompsoc
  % The Computer Society usually uses the plural form
  \section*{Acknowledgments}
\else
  % regular IEEE prefers the singular form
  \section*{Acknowledgment}
\fi
J. Navarro acknowledges support from Ministero de Econom{\'i}a y Competitividad of the Spanish Government under grant TIN2017-85572-P (MINECO/AEI/FEDER, UE).

% Can use something like this to put references on a page
% by themselves when using endfloat and the captionsoff option.
\ifCLASSOPTIONcaptionsoff
  \newpage
\fi

% trigger a \newpage just before the given reference
% number - used to balance the columns on the last page
% adjust value as needed - may need to be readjusted if
% the document is modified later
%\IEEEtriggeratref{8}
% The "triggered" command can be changed if desired:
%\IEEEtriggercmd{\enlargethispage{-5in}}

% references section

% can use a bibliography generated by BibTeX as a .bbl file
% BibTeX documentation can be easily obtained at:
% http://mirror.ctan.org/biblio/bibtex/contrib/doc/
% The IEEEtran BibTeX style support page is at:
% http://www.michaelshell.org/tex/ieeetran/bibtex/
%\bibliographystyle{IEEEtran}
% argument is your BibTeX string definitions and bibliography database(s)
%\bibliography{IEEEabrv,../bib/paper}
%
\bibliographystyle{IEEEtran}
\bibliography{refs}

% biography section
% 
% If you have an EPS/PDF photo (graphicx package needed) extra braces are
% needed around the contents of the optional argument to biography to prevent
% the LaTeX parser from getting confused when it sees the complicated
% \includegraphics command within an optional argument. (You could create
% your own custom macro containing the \includegraphics command to make things
% simpler here.)
%\begin{IEEEbiography}[{\includegraphics[width=1in,height=1.25in,clip,keepaspectratio]{mshell}}]{Michael Shell}
% or if you just want to reserve a space for a photo:

\vfill

\begin{IEEEbiographynophoto}{Julia Navarro}
received the BSc in Mathematics from the Universitat de les Illes Balears, Spain, in 2014 and the MSc in Computer Vision from the Universitat Aut{\`o}noma de Barcelona, Spain, in 2015. She is currently pursuing the PhD degree with the Universitat de les Illes Balears. Her research interests are computer vision and mathematical image analysis. 
\end{IEEEbiographynophoto}

\begin{IEEEbiographynophoto}{Neus Sabater}
received the BSc degree in 2005 from the Universitat de Barcelona, Spain, and the MSc and PhD degrees in 2006 and 2009, respectively, in image processing from the Ecole Normale Supérieure de Cachan, France. She was a postdoctoral researcher at the California Institute of Technology before being appointed at Technicolor Research \& Innovation in 2011 where she is currently a Senior Scientist. Her research interest include image processing, computer vision and computational photography.
\end{IEEEbiographynophoto}

%\begin{IEEEbiography}{Michael Shell}
%Biography text here.
%\end{IEEEbiography}
%
%% if you will not have a photo at all:
%\begin{IEEEbiographynophoto}{John Doe}
%Biography text here.
%\end{IEEEbiographynophoto}
%
%% insert where needed to balance the two columns on the last page with
%% biographies
%%\newpage
%
%\begin{IEEEbiographynophoto}{Jane Doe}
%Biography text here.
%\end{IEEEbiographynophoto}

% You can push biographies down or up by placing
% a \vfill before or after them. The appropriate
% use of \vfill depends on what kind of text is
% on the last page and whether or not the columns
% are being equalized.

%\vfill

% Can be used to pull up biographies so that the bottom of the last one
% is flush with the other column.
%\enlargethispage{-5in}

% that's all folks
\end{document}